\documentclass[final]{cvpr}

\usepackage{times}
\usepackage{epsfig}
\usepackage{graphicx}
\usepackage{amsmath}
\usepackage{amssymb}
\usepackage[noend]{algpseudocode}
\usepackage{algorithm}
\usepackage{color}
\usepackage{subcaption}
\usepackage[table,dvipsnames]{xcolor}
\usepackage{multirow}
\usepackage{makecell}
\usepackage{tabularx}
\usepackage{placeins}

\usepackage{xspace}
\usepackage{dsfont}

\usepackage{mathtools,xparse}
\usepackage{siunitx}
\usepackage{stmaryrd}
\usepackage{mdframed}
\usepackage{booktabs}
\usepackage{fontawesome}
\usepackage{colortbl}
\usepackage{amssymb}

\usepackage[pagebackref=true,breaklinks=true,colorlinks,bookmarks=false]{hyperref}

\def\cvprPaperID{2531} % *** Enter the CVPR Paper ID here
\def\confYear{CVPR 2021}

\newcommand*{\SE}{\text{SE(3)}}
\newcommand*{\AStar}{\text{A}^*}
\newcommand*{\Edges}{\mathcal{E}}
\newcommand*{\Walk}{\mathcal{W}}
\newcommand*{\Vertices}{\mathcal{V}}
\newcommand*{\Graph}{\mathcal{G}}
\renewcommand*{\eg}{\emph{e.g.}\@\xspace}
\renewcommand*{\ie}{\emph{i.e.}\@\xspace}

\newcommand{\mat}[1]{\mathbf{{#1}}}
\definecolor{wincolor}{rgb}{0.95, 0.2, 0.2}

\robustify\bfseries % bold in table with this
\newcolumntype{Y}{>{\centering\arraybackslash}X}

\begin{document}

%%%%%%%%% TITLE
\title{Efficient Initial Pose-graph Generation for Global SfM}

\author{Daniel Barath$^{1,2}$, Dmytro Mishkin$^{1}$, Ivan Eichhardt$^{2}$, Ilia Shipachev$^{1}$, and Jiri Matas$^{1}$\\
$^1$ Visual Recognition Group, Faculty of Electrical Engineering, 
Czech Technical University in Prague \\
$^2$ Machine Perception Research Laboratory, 
SZTAKI, Budapest \\
{\tt\small barath.daniel@sztaki.mta.hu}
}
\maketitle
%\thispagestyle{empty}

%%%%%%%%% ABSTRACT
\begin{abstract}
   We propose ways to speed up the initial pose-graph generation for global Structure-from-Motion algorithms. 
   To avoid forming tentative point correspondences by FLANN and geometric verification by RANSAC, which are the most time-consuming steps of the pose-graph creation, we propose two new methods -- 
   built on the fact that image pairs usually are matched consecutively. 
   Thus, candidate relative poses can be recovered from paths in the partly-built pose-graph.
   We propose a heuristic for the $\AStar$ traversal, considering global similarity of images and the quality of the pose-graph edges. 
   Given a relative pose from a path, descriptor-based feature matching is made ``light-weight'' by exploiting the known epipolar geometry.
   To speed up PROSAC-based sampling when RANSAC is applied, we propose a third method
   to order the correspondences by their inlier probabilities from previous estimations. 
   The algorithms are tested on \num{402130} image pairs from the 1DSfM dataset and they speed up the feature matching \num{17} times and pose estimation \num{5} times.
   %The source code will be made public.
\end{abstract}

%%%%%%%%% BODY TEXT
\section{Introduction}
\label{sec:intro}
% Topic: global SfM
Structure-from-Motion (SfM) has been intensively researched in computer vision for decades. 
Most of the early methods adopt an incremental strategy, where the reconstruction is built progressively and the images are carefully added one-by-one in the procedure~\cite{rother2003multi, pollefeys2004visual, pollefeys2008detailed, agarwal2011building, wu2013towards, schonberger2016structure}.
Recent studies~\cite{govindu2001combining, govindu2004lie, brand2004spectral, martinec2007robust, arie2012global, chatterjee2013efficient, hartley2013rotation, ozyesil2015robust, cui2015global, carlone2015initialization, zhu2018very} show that global approaches, considering all images simultaneously when reconstructing the scene, lead to comparable or better accuracy than incremental techniques while being significantly more efficient. Also, global methods are less dependent on local decisions or image ordering. 

%\todo{Dmytro: We should say smth about disadvantages of the global SfMs here. Speed? }
% \todo{Incremental are as good as global if not etter, but ofcc in incremental its critial in which way you sequence the increment. We show that starting sort of global...}

\begin{figure}
  	\centering
  	\includegraphics[width=1.0\columnwidth]{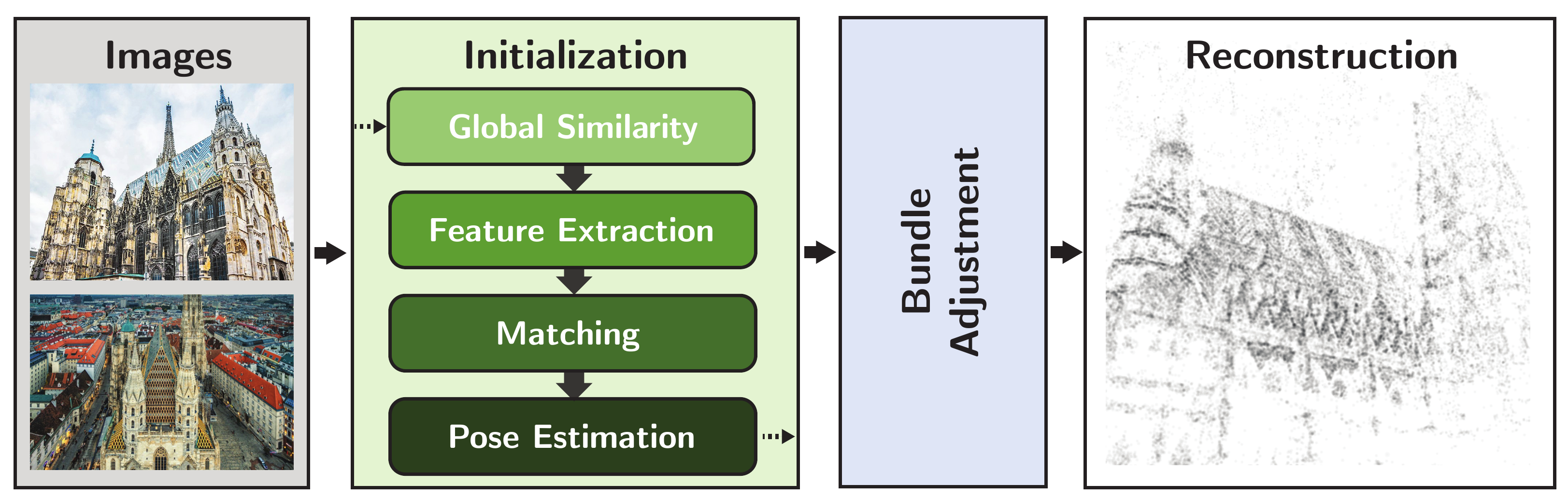}
    \caption{ The architecture of a global SfM pipeline. }
    \label{fig:globalsfm}
\end{figure}

% How global SfM usually proceeds
Typically, structure-from-motion pipelines consist of the following steps, see Fig.~\ref{fig:globalsfm}. 
First, features are extracted in all images. Such step is easily parallelizable and has $\mathcal{O}(n)$ time complexity, where $n$ is the number of images to be included in the reconstruction. 
These features are then often used to order the image pairs from the most probable to match to the most difficult ones, {\eg}, via bag-of-visual-words~\cite{VideoGoogle2003}.
Next, tentative correspondences are generated between all image pairs by matching the often high-dimensional (\eg, $128$ for SIFT~\cite{SIFT2004}) descriptors of the detected feature points. 
Then, the correspondences are filtered and relative poses are estimated between all image pairs by applying RANSAC~\cite{fischler1981random}. 
Usually, the feature matching and geometric estimation steps are by far the slowest parts, both having quadratic complexity in the number of images. Moreover, feature matching has a quadratic worst-case time complexity as it depends on the product of the number of features in the respective images.
Finally, a global bundle adjustment obtains the accurate reconstruction from the pair-wise poses. Interestingly, this step has negligible time demand, \ie, a few minutes in our experiments, compared to the initial pose-graph generation.

\begin{figure}
  	\centering
  	\fcolorbox{gray}{white}{
  	\includegraphics[width=0.44\columnwidth]{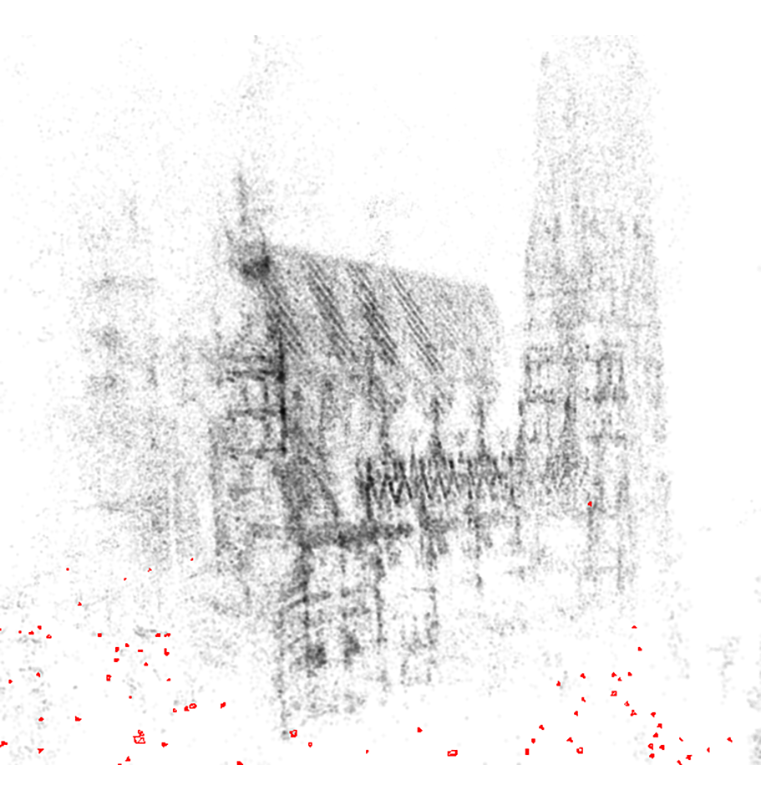} }
  	\fcolorbox{gray}{white}{
  	\includegraphics[width=0.44\columnwidth]{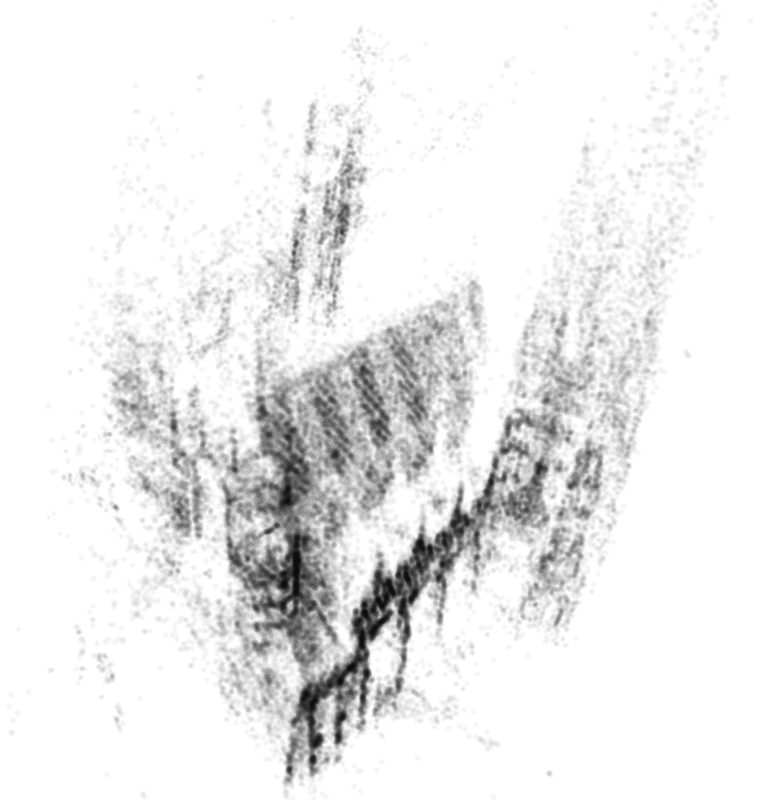}
  	}
    \caption{ Reconstruction by initializing Theia's~\cite{theia-manual} global SfM with the pose-graph from the proposed algorithms. }
    \label{fig:teaser}
\end{figure}

% Our solution
%\todo{Swap RANSAC and tentative correspondences.}
This paper has three major contributions -- three new algorithms which allow removing the need of RANSAC-based geometric estimation and, also, to make descriptor-based feature matching ``light-weight''.
\textit{First}, a method is proposed exploiting the partly-built pose-graph to avoid the computationally demanding RANSAC-based robust estimation. To do so, we propose a heuristic for the $\AStar$~\cite{AStar1968} algorithm which guides the path-finding even without having a metric distance between the views. 
The lack of such a distance originates from the fact that the edges of a pose-graph represent relative poses and, thus, neither the global scale nor the length of any of the translations are known. 
\textit{Second}, we propose a technique to make the expensive descriptor-based feature matching ``light-weight'' by using the pose determined by $\AStar$.
This guided matching approach uses the fundamental matrix to efficiently select keypoints, which lead to correspondences consistent with the pose, via hashing.
\textit{Third}, an algorithm is proposed to adaptively re-rank the point-to-point correspondences based on their history -- whether one or both of the points had been inliers in previous estimations.
The method exploits the fact that these inlying feature points likely represent 3D points consistent with the rigid reconstruction of the scene.
This adaptive ranking speeds up the robust estimation by guiding PROSAC~\cite{chum2005matching} to find a good sample early.
The proposed techniques were tested on the 1DSfM dataset~\cite{wilson_eccv2014_1dsfm}, see Fig.~\ref{fig:teaser} for an example reconstruction. They consistently and significantly speed up the pose-graph generation.

%\todo{We present a method which allows removing the need of RANSAC and descriptor-based feature matching.}
%In this paper, we focus on speeding up both the RANSAC and feature matching procedures by exploiting the fact that the image pairs are usually matched consecutively on a single or multiple CPU cores. 
%\todo{The only thing which remains quadratic is the global similarity}

\subsection{Related Techniques}
\label{sec:notation}

\noindent
\textbf{Robust estimation.}
To speed up robust estimation, there has been a number of algorithms proposed over the years. 
NAPSAC~\cite{nasuto2002napsac}, PROSAC~\cite{chum2005matching} and P-NAPSAC~\cite{barath2020magsacpp} modify the RANSAC sampling strategy to increase the probability of selecting an all-inlier sample early. 
PROSAC exploits an a priori predicted inlier probability rank of the points and starts the sampling with the most promising ones. NAPSAC is built on the fact the real-world data often are spatially coherent and selects samples from local neighborhoods, where the inlier ratio is likely high.
P-NAPSAC combines the benefits of PROSAC and NAPSAC by first sampling locally, and then progressively blending into global sampling.
The Sequential Probability Ratio Test~\cite{chum2008optimal} (SPRT), inspired by Wald's theory, is applied for rejecting models early if the probability of being better than the previous best model falls below a threshold. 
All of the mentioned RANSAC improvements consider the case of a single, isolated two-view robust estimation. Here, we exploit information arising while performing estimation on some subset of the $\binom{N}{2}$ image pairs where some images are matched more than once.

\noindent
\textbf{Feature matching} can be sped up in several ways, \eg, by the use of  binary descriptors~\cite{ORB2011,FREAK2012,BinBoost2015} 
or by limiting the number of features detected, as often done in SLAM systems~\cite{Mur15}. However, this often results in inaccurate camera poses for the general 3D reconstruction problem~\cite{IMC2020}. 
Often, approximate nearest neighbor algorithms are employed, such as kd-tree or product quantization~\cite{FLANN2009, PQ2011}. 
Hardware-based speed-ups include using a GPU~\cite{FAISS2017}.
None of these techniques consider that the matching is performed on a number of image pairs, where the relative pose might be known, at least approximately, prior to the matching.

\noindent
\textbf{Global image similarity.}
Matching an \textit{unordered} image collection is usually a harder and more time consuming task than, matching, {\eg}, a video sequence. There are two reasons for that. First, many image pairs might not have any commonly visible part of the scene and the time spent on matching attempts is wasted. Moreover, no-match is the worst case scenario for RANSAC, which will run the maximum number of iterations, often orders of magnitude more than in the matching-possible case.
Second, the time spent on the estimation of epipolar geometry highly depends on the inlier ratio of the tentative correspondences~\cite{magsac2019}. The inlier ratio, in turn, depends on the difference between the two viewpoints: the bigger the difference, the fewer tentative correspondences are correct~\cite{MikoDetEval2005, MikoDescEval2003}.
A natural question would be -- is it possible to order the image pairs from the most probable ones to the most difficult or impossible to match? Image retrieval techniques are commonly used for it, {\eg}, one could re-use extracted local features to find the most promising candidates for matching via bag-of-visual-words~\cite{VideoGoogle2003} and then quickly reorder the preliminary list using geometric constraints~\cite{Philbin07, schoenberger2016vote} as it is implemented in COLMAP SfM~\cite{schonberger2016structure}. 
Such systems work well, but have significant memory footprint and are now overcome by CNN-based global descriptors~\cite{GeM2018, RevOxford2018, GLDv2_2020}, which are both faster to compute and provide more accurate results. 

We use the following approach to generate a fully connected image similarity graph as a preliminary step. First, we extract GeM~\cite{GeM2018} descriptors with ResNet-50~\cite{He2016ResNet} CNN, pre-trained on GLD-v1 dataset~\cite{DELF2017}. Then we calculate the inner-product similarity between all the descriptors, resulting in an $n\times n$ similarity matrix. The calculation of the similarity matrix is the only quadratic step of our pipeline. However, the scalar product operation is extremely fast. In practice, the creation and processing of the similarity matrix takes negligible time. 

% DM the following is not true yet. I have implemented it later, and we need to check if it works better
% Finally, we refine the similarity matrix by running the diffusion-based re-ranking~\cite{Diffusion2017}.

%%%%%%%%%%%%%%%%
\begin{figure}
  	\centering
  	\includegraphics[width=0.75\columnwidth]{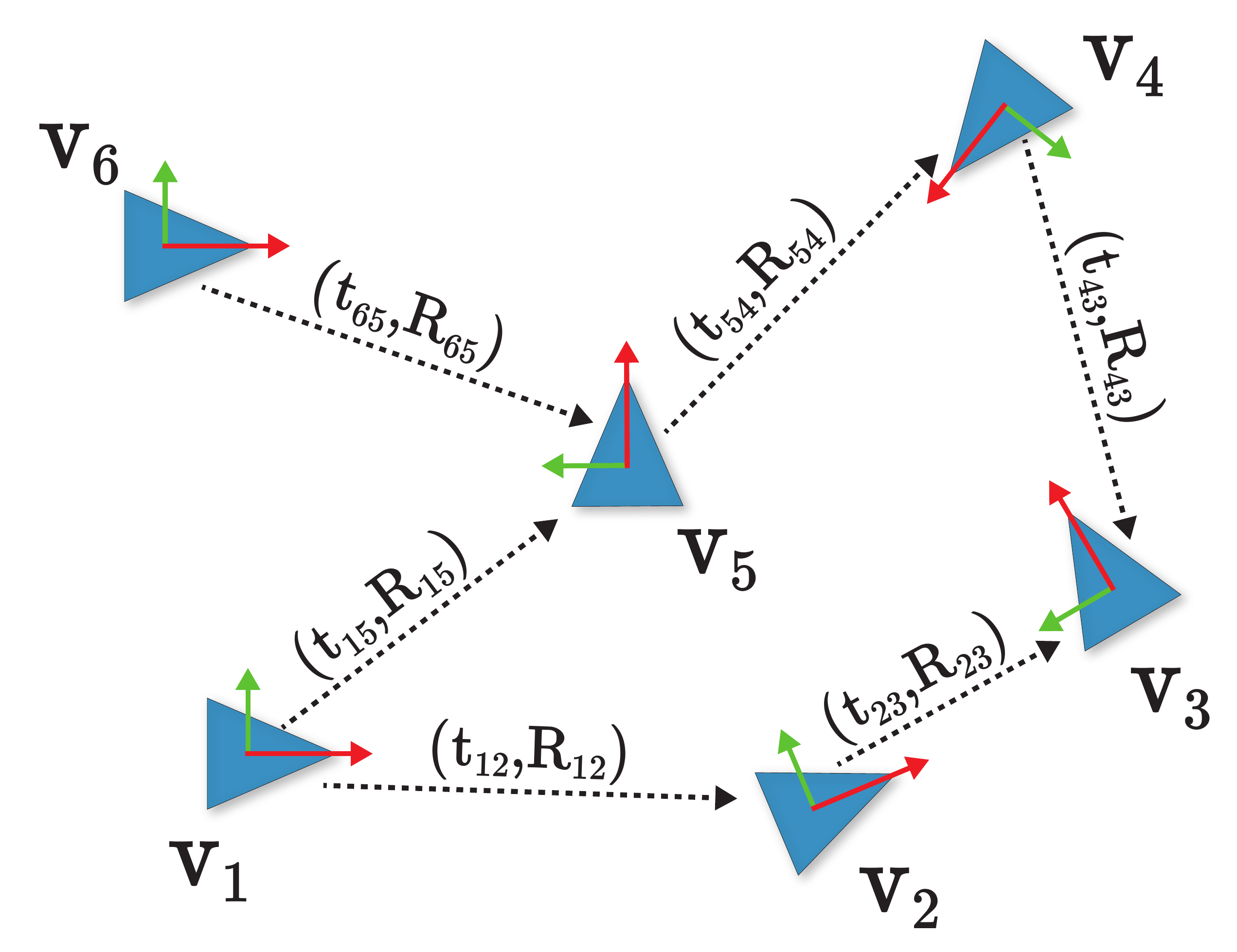}
    \caption{ A schematic pose-graph, used for initializing global SfM algorithms. Vertices (images) are connected by edges representing relative pose $(\mat t_{ij}, \mat R_{ij}) \in \SE$.}
    \label{fig:posegraph}
\end{figure}

\begin{figure*}
  	\centering
  %	\begin{subfigure}[t]{1.0\textwidth}
      	\includegraphics[width=0.32\textwidth]{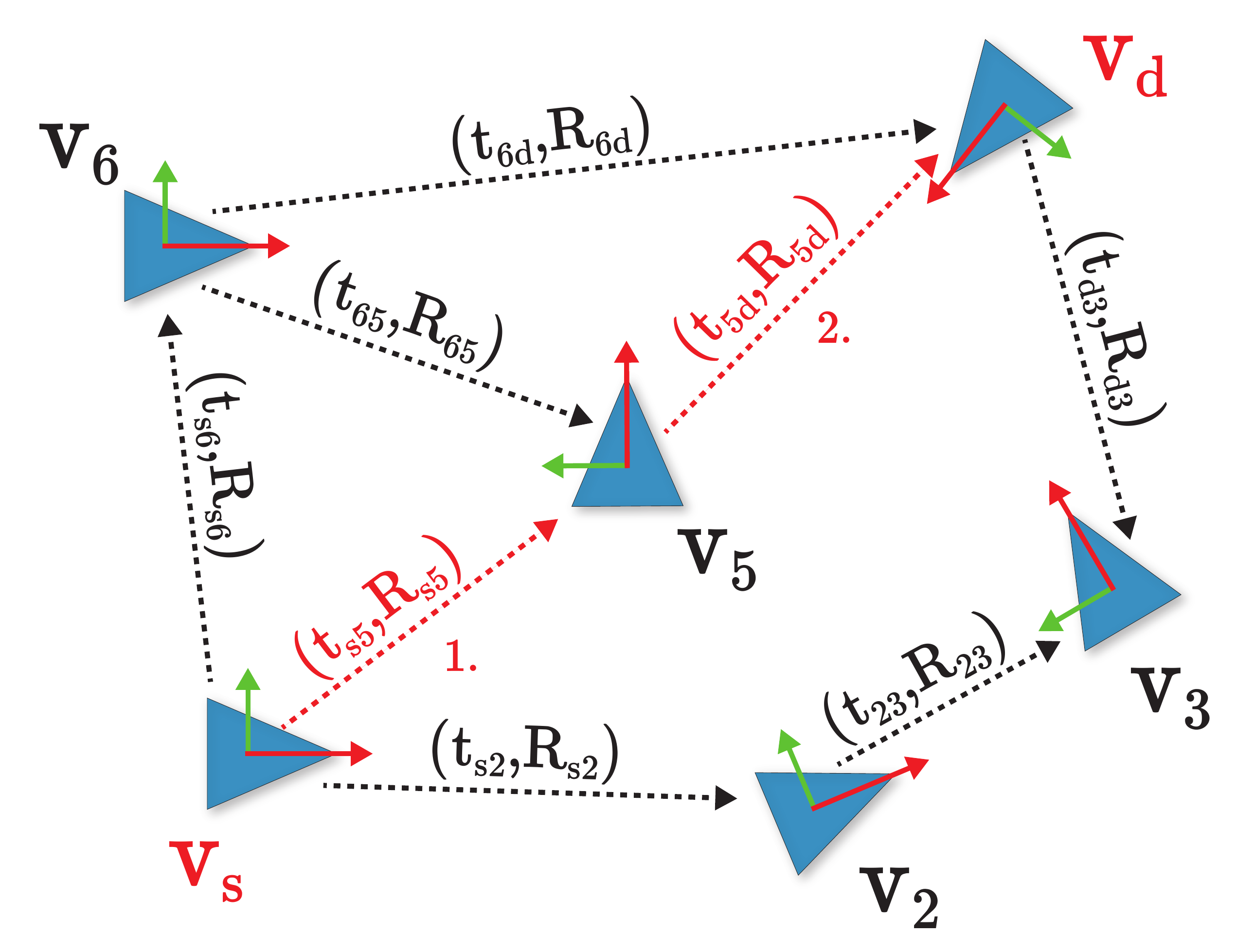}
      	\includegraphics[width=0.32\textwidth]{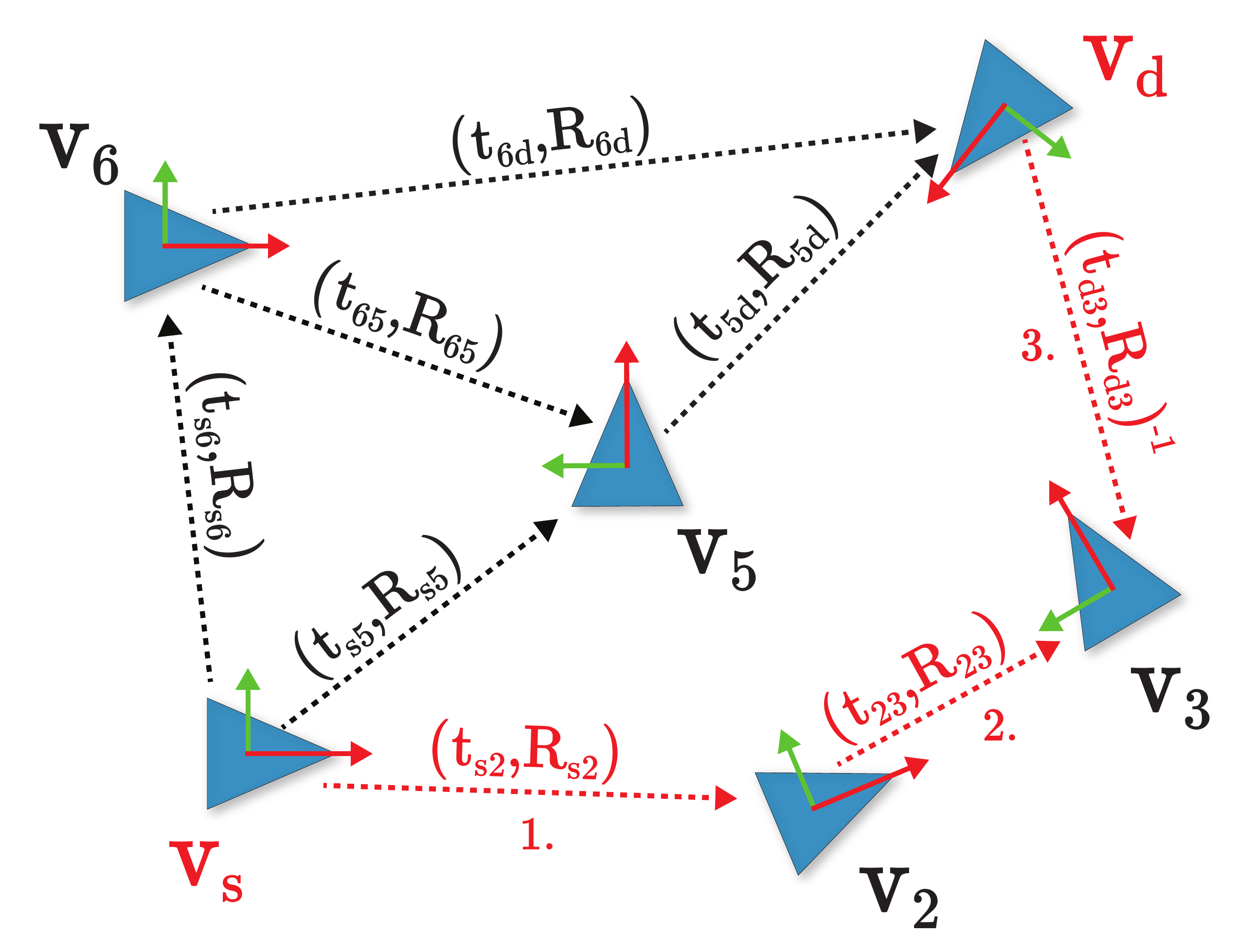}
      	\includegraphics[width=0.32\textwidth]{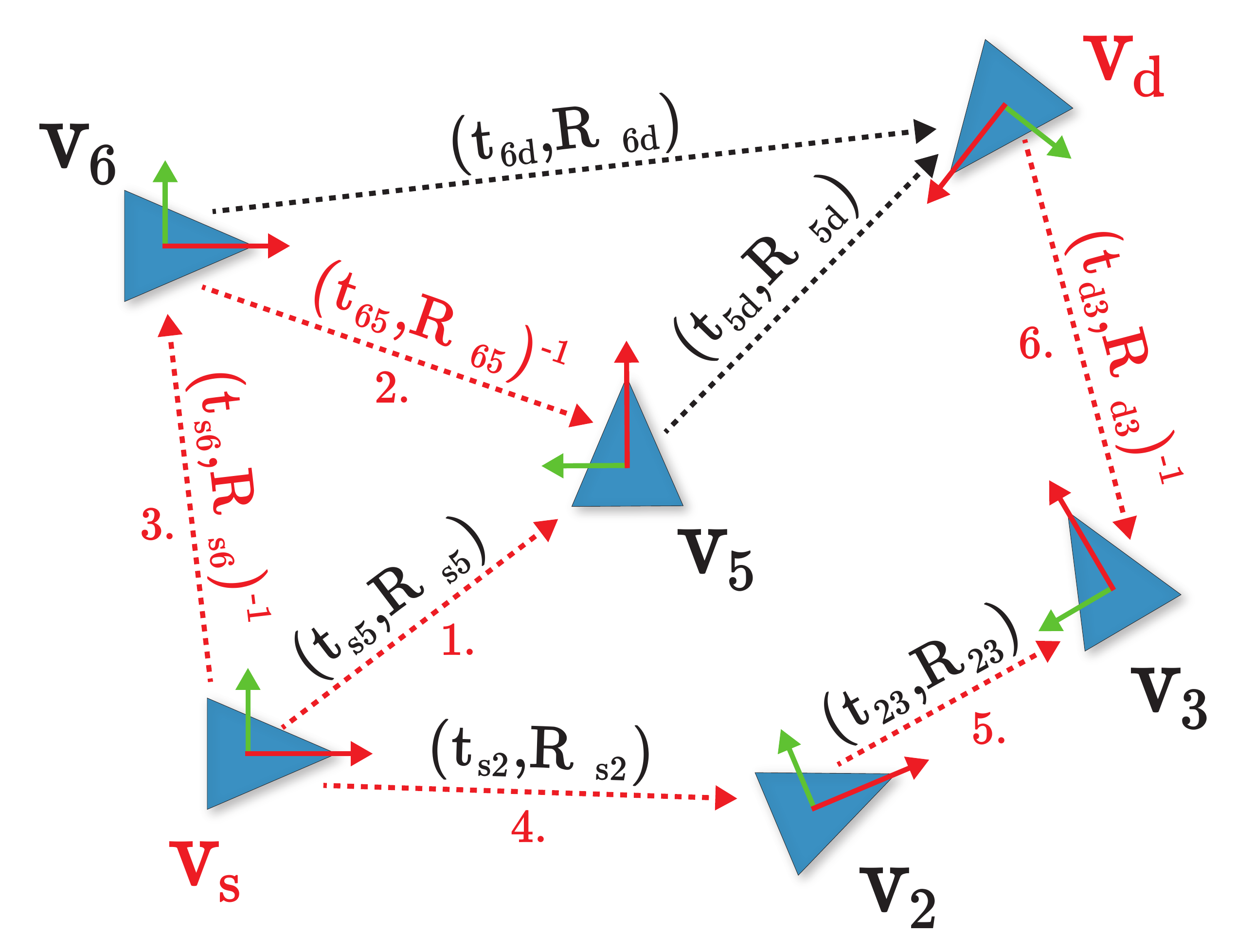}
  %	    \caption{Without edge inversion allowed}
  %  \end{subfigure}
  %	\begin{subfigure}[t]{1.0\textwidth}
  %    	\includegraphics[width=0.32\textwidth]{assets/walk_1i.pdf}
  %    	\includegraphics[width=0.32\textwidth]{assets/walk_3i.pdf}
  %    	\includegraphics[width=0.32\textwidth]{assets/walk_2i.pdf}
 % 	    \caption{With edge inversion allowed}
 %   \end{subfigure}
    \caption{ Example walks between vertices $v_s$ and $v_d$ with edge inversion allowed. The relative pose $\mat P_{sd} = (\mat t_{sd}, \mat R_{sd}) \in \SE$ from $v_s$ to $v_d$ is calculated as $\mat P_{sd} = \mat P_{nd} \dots \mat P_{12} \mat P_{s1}$, where $\mat P_{s1}$ is the pose of the first edge in the walk and $\mat P_{nd}$ is that of the last one, $n$ is the length of the walk.  }
    \label{fig:walks}
\end{figure*}

\begin{table}
	\center
	\resizebox{0.999\columnwidth}{!}{\begin{tabular}{ c l }
    \hline
   		\multicolumn{2}{ c }{ \cellcolor{black!10}Symbols used in this paper\rule{0pt}{2.0ex} } \\[0.2mm]
    \hline 
   		$\Graph = (\Vertices, \Edges)$ & - Directed graph, \\
   		$v \in \Vertices$ & - A vertex from the vertex set \\ 
   		$e = (v_i, v_j) \in \Edges$ & - Edge between vertices $v_i$ and $v_j$ \\
   		$f \in \{ e, e^{-1} \}$ & - An edge or its inverse \\
   		%$\SE$ & - Group of 6D poses \\ 
   		$\phi(e) : \Edges \to \SE$ & - Relative pose of edge $e$ \\ 
   		$\rho(e) : \Edges \to \mathbb{R}$ & - Quality of edge $e$ \\ 
   		$\delta(v_i, v_j) : \Vertices \times \Vertices \to \mathbb{R}$ & - Distance of vertices $v_i$ and $v_j$ \\ 
   		$\Walk \in \{ (f_1, \dots, f_{n}) \; | \; 1 < n \}$ & - A walk \\ 
   		$\rho(\Walk) : \{ (f_1, \dots, f_{n})\} \to \mathbb{R}$ & - Quality of walk $\Walk$  \\ 
    \hline     
\end{tabular}}
\label{tab:notation}
\end{table}
    
\section{Relative Pose from Directed Walks}
\label{sec:rel-pose-from-walks}
We propose an approach to speed up the pose-graph generation by avoiding running RANSAC when possible. 
The core idea exploits the fact that when estimating the relative pose between the $(t+1)$th image pair from an image collection, we are given a pose-graph consisting of $t$ edges, \ie, $t$ view pairs. 
This pose-graph can often be used to estimate the pose without running RANSAC-like robust estimation. 

In the rest of the description, we assume that the view pairs are ordered by their similarity scores. 
Thus, we start the pose estimation from the most similar view pair.
Let us assume that we have already matched $t$ image pairs successfully and, thus, we are given pose-graph $\Graph_t = (\Vertices, \Edges_{t})$, where $\Edges_t \subseteq \{ (v_1, v_2) \; | \; v_1, v_2 \in \Vertices \}$ are the $|\Edges_t| = t$ edges and $\Vertices$ is the set of images in the dataset, see Fig.~\ref{fig:posegraph} for an example.
Function $\phi : \Edges_t \to \SE$ maps edge $e \in \Edges_t$ to its estimated relative pose. 
%For example, given $e = (v_1, v_2)$, the relative rotation and translation between views $v_1$ and $v_2$ is $\phi(e)$, 

When estimating the relative pose between the $(t+1)$th view pair, we are given two options. 
The traditional one is to run robust estimation on the corresponding points between the two images. The estimated pose $\mat P \in \SE$ is then added to the pose-graph as the pose of the new edge. Thus, $\Edges_{t+1} = \Edges_{t} \cup \{ e = (v_s, v_d) \}$\footnote{$v_s$ -- source view, $v_d$ -- destination view} and $\phi(e) = \mat P$.
The problem with this step is that when having few inliers and, thus, low inlier ratio, the estimation can be often time-consuming.
Due to this step being done approximately $\binom{|\Vertices|}{2}$ times, the slow pair-wise pose estimation has a \textit{severe impact} on the processing time of the entire pose-graph estimation. 

Therefore, instead of estimating the pose blindly between a pair of views $(v_s, v_d)$, we propose to use the previously generated pose-graph $\Graph_t$. 
Let us assume that there exists a finite directed walk $\mathcal{W} = (f_{w_1}, f_{w_2}, \dots, f_{w_{n - 1}})$, for which there is a sequence of vertices $(v_{w_1}, v_{w_2}, \dots, v_{w_n})$ such that $f_{w_i} \in \{ e_{w_i}, e_{w_i}^{-1} \}$, $e_{w_i} = (v_{w_i}, v_{w_{i + 1}})$ for $i = 1, 2, \dots, n - 1$, and $v_{w_1} = v_s$, $v_{w_n} = v_d$. See Fig.~\ref{fig:walks} for examples.
The direction of edge $e$ can be inverted as $e^{-1}$ by inverting the relative pose as $\phi(e^{-1}_i) = \phi(e_i)^{-1}$ and swapping its vertices as $e^{-1}_i = (v_{i + 1}, v_i)$. 
We define the pose implied by walk $\Walk$ recursively as 
\begin{equation}
    \begin{array}{l}
        \phi(\Walk) \\
        = \phi(f_{w_1}, f_{w_2}, \dots, f_{w_{n - 1}}) \\
        = \phi(f_{w_1}, f_{w_2}, \dots,f_{w_{n - 2}}) \phi(f_{w_{n - 1}}) \\
        = \phi(f_{w_1}, f_{w_2}, \dots,f_{w_{n - 3}}) \phi(f_{w_{n - 2}}) \phi(f_{w_{n - 1}}) \\
        = \dots \\
        = \phi(f_{w_1}) \phi(f_{w_2}) \dots \phi(f_{w_{n - 1}}).
    \end{array}
    \label{eq:walkpose}
\end{equation}
Consequently, the relative pose between views $v_s$ and $v_d$ is calculated as $\phi(\Walk)$ given a finite walk $\Walk$.

The problem with \eqref{eq:walkpose} is that a single incorrectly estimated pose $\phi(f)$, $f \in \Walk$, makes the entire $\phi(\Walk)$ wrong.
Therefore, we aim at finding \textit{multiple walks} within a given distance, \ie, the maximum depth is restricted to avoid infinitely long walks.
The walks returned are evaluated sequentially and immediately, see  Alg.~\ref{alg:main_alg}. 
Whenever a new walk $\Walk$ is found, its inlier ratio is calculated from pose $\phi(\Walk)$ and the correspondences between the source and destination images, $v_s$ and $v_d$, respectively. 

\noindent
\textbf{Termination.} 
There are two cases when the procedure of finding and testing walks terminates. 
They are as follows: \\[1mm]
1.~The process finishes when there are no more walks found within the maximum distance.\\[1mm] 
2.~If there is a reasonably good pose $\mat{P}$ found, the process terminates. We consider a relative pose reasonably good if it has at least $I_{\text{min}}$ inliers.\footnote{
Parameter $I_{\text{min}}$ is typically set to \num{20} in most of the recent Structure-from-Motion algorithms~\cite{theia-manual,schonberger2016structure}.
}
\\[1mm]
\noindent
\textbf{Pose refinement.}
In case the pose is obtained successfully from one of the walks, it is calculated solely from the edges of pose-graph $\Graph_t$ without considering the correspondences between images $v_s$ and $v_d$.
In order to improve the accuracy and obtain $\mat{P}^*$, we apply iteratively re-weighted least-squares fitting initialized by the newly estimated model $\mat{P}$.
Finally, $\Edges_{t+1} = \Edges_{t} \cup \{ e = (v_s, v_d) \}$ and $\phi(e) = \mat P^*$.

\noindent
\textbf{Failures.}
There are cases when at least a single walk exists between views $v_s$ and $v_d$, but the implied pose is incorrect, \ie, it does not lead to a reasonable number of inliers. 
In those cases, we apply the traditional approach, \ie, RANSAC-based robust estimation. 
%We use Graph-Cut RANSAC~\cite{barath2018graph} for robust estimation.

\noindent
\textbf{Visibility.}
Deciding if there is at least a single walk in the pose-graph between views $v_s$ and $v_d$ can be done by the union-find algorithm in $\mathcal{O}(1)$ time.
%For this purpose, a hash-map with a suitable hashing function is updated every
On average, the time complexity of the update is $\mathcal{O}(\log(n))$.

\noindent
\textbf{Parallel graph building.}
Building graph $\Graph$, checking the visibility, finding and evaluating walks in parallel on multiple CPUs is efficiently doable by using readers-writer locking mechanisms where each thread matches the next best view pair. The readers are the processes trying to find walks between two views or the ones checking if view $v_s$ is visible from $v_d$. A process becomes writer only when it adds a new edge to the pose-graph or updates the union-find method for visibility checking which both takes only a few operations.

\begin{algorithm}[t]
\begin{algorithmic}[1]
	\Statex{\hspace{-1.0em}\textbf{Input:} $\Graph_t$ -- current pose-graph; $v_s$, $v_d$ -- views to match}
	\Statex{\hspace{-1.0em}\phantom{\textbf{Input:}} $V$ -- visibility table; $\mathcal{P}$ -- point correspondences }
	\Statex{\hspace{-1.0em}\phantom{\textbf{Input:}} $d_{\text{max}}$ -- maximum depth }
    \Statex{\hspace{-1.0em}\textbf{Output:} $\mat P$ -- pose; $\mathcal{I}$ -- inliers}
   	\Statex{}
    \If{$\neg\texttt{Visible}(V, v_s, v_d)$}
        \State{\textbf{return}}
    \EndIf
    \While{$\neg \; \texttt{Terminate}(\mathcal{I})$}
    	\State{$\Walk \leftarrow \texttt{GetNextWalk}(\Graph_t, v_s, v_d, d_{\text{max}})$}
    	\If{$\texttt{EmptyWalk}(\Walk)$}
    	    \State{\textbf{break}}
    	\EndIf
    	\State{$\mat P_\Walk \leftarrow \phi(\Walk)$}
    	\State{$\mathcal{I}_\Walk \leftarrow \texttt{GetInliers}(\mat P_\Walk, \mathcal{P})$}
    	\If{$|\mathcal{I}_\Walk| > |\mathcal{I}|$}
    	    \State{$\mat P \leftarrow \mat P_\Walk, \mathcal{I} \leftarrow \mathcal{I}_\Walk$}
    	\EndIf
    \EndWhile
\end{algorithmic}
\caption{\bf Pose from Pose-Graph.}
\label{alg:main_alg}
\end{algorithm}

\begin{algorithm}
\begin{algorithmic}[1]
	\Statex{\hspace{-1.0em}\textbf{Input:} $\mathcal{K}_1$, $\mathcal{K}_2$ --  sets of keypoints; $\mat R, \mat t$ -- relative pose;}
	\Statex{\hspace{-1.0em}\phantom{\textbf{Input:}} $\mu$ -- inlier-outlier threshold; $b$ -- bin number}
    \Statex{\hspace{-1.0em}\textbf{Output:} $\mathcal{P}$ -- point correspondences}
   	\Statex{}
    \State{$\mat E = [\mat t]_\times \mat R$}\Comment{Get essential matrix}
    \State{$\mathcal{B} \leftarrow \texttt{Hashing}(\mathcal{K}_2, \mat E, b)$}
    \For{$\mat p_1 \in \mathcal{K}_1$}
        \State{$\mat d_1 \leftarrow \texttt{Descriptor}(\mat p_1), \delta \leftarrow \infty, \mat p^* \leftarrow \mat 0$}
        \For{$\mat p_2 \in \mathcal{B}(\mat p_1)$}
            \If{ $\epsilon(\mat p_1, \mat p_2, \mat E) < \mu$ } \Comment{Sampson dist.}
                \State{$\mat d_2 \leftarrow \texttt{Descriptor}(\mat p_2)$}
                \If{ $|\mat d_2 - \mat d_1| < \delta$ }
                    \State{$\delta \leftarrow |\mat d_2 - \mat d_1|, \mat p^* \leftarrow \mat p_2$}
                \EndIf
            \EndIf
        \EndFor
        \State{$\mathcal{P} \leftarrow \mathcal{P} \cup \{ (\mat p_1, \mat p^*) \}$}
    \EndFor
\end{algorithmic}
\caption{\bf Epipolar Hashing.}
\label{alg:main_alg_on_demand}
\end{algorithm}

\subsection{Pose-graph Traversal}

It is a rather important question how to find a walk between views $v_s$ and $v_d$ efficiently. 
There are a number of graph traversals, however, most of them are not suitable for returning walks in a large graph in reasonable time. 
We choose the $\AStar$~\cite{AStar1968} algorithm since it works well for such a task when a good heuristic exists.
In this section, we propose a way of obtaining multiple walks in pose-graph $\Graph_t$ by defining a heuristic for the $\AStar$ algorithm.

%The depth-first algorithm likely falls to infinite cycles even in small graphs when looking for walks. Even with a depth limit, its worst-case run-time is far unacceptable for such a task.
%The breadth-first traversal is a good choice to surely find all walks within a given depth, however, it becomes extremely slow for big graphs.
%Even though the bidirectional breadth-first algorithm is notably faster, the same holds for big graphs. 
% For such a path-finding problem, usually, the $\AStar$~\cite{AStar1968} algorithm is a suitable choice if a good heuristic exists.

%\subsection{Heuristic for \textbf{$\AStar$}} 

The objective is to define a heuristic which guides the $\AStar$ algorithm from node $v_s$ to node $v_d$ while visiting as few vertices as possible.
Since we are given a graph of relative poses, we are unable to define a metric, measuring the Euclidean distance of a view pair. When having relative poses, both the global and local scales remain unknown and, thus, all translations have unit length. As a consequence, it is unclear whether two views are close to or far from each other. 
The proposed heuristic is composed of two functions.

First, the global similarity of views $v_s$ and $v_d$ are measured as $\delta(v_s, v_d)$, $\delta : \Vertices \times \Vertices \to \mathbb{R}$.
It is determined via the inner-product of GeM~\cite{GeM2018} descriptors with ResNet-50~\cite{He2016ResNet} CNN, pre-trained on GLD-v1 dataset~\cite{DELF2017} as described earlier.
Second, reflecting the fact that a single incorrectly estimated edge severely affects the pose of the entire walk, we also consider the quality of edge $e$ via function $\rho(e) : \Edges \to \mathbb{R}$. To our experiments, the inlier ratio is usually a good indicator of the pose quality. Function $\rho(e)$ returns the inlier ratio calculated given the pose $\phi(e)$ of the current edge and the point correspondences. 

To measure the quality of the entire walk $\Walk$, we have to consider that a single incorrect pose makes $\phi(W)$ incorrect as well. Thus, the quality of $\Walk$ is measured as $Q(\Walk) = \min_{f \in \Walk} \rho(f)$, \ie, the quality of the least accurate edge.
To measure the similarity of walk $\Walk$ between the destination view $v_d$, we define function $\Delta(\Walk, v_d) = \max_{f = (v_1, v_2) \in \Walk} \delta(v_2, v_d)$,
\ie, the most similar vertex determines the similarity.
The heuristic considering both the quality of the walk and similarity to the destination is as
\begin{equation}
    h(\Walk) = \lambda \min_{f \in \Walk} \rho(f) + (1 - \lambda) \max_{f = (v_1, v_2) \in \Walk} \delta(v_2, v_d),
    \label{eq:heuristics}
\end{equation}
where $\lambda \in [0, 1]$ is a weighting parameter.
Expression $\min_{f \in \Walk} \rho(f)$ forces the $\AStar$ algorithm to find a walk maximizing the minimum inlier ratio along the walk. 
Expression $\max_{f = (v_1, v_2) \in \Walk} \delta(v_2, v_d)$ affects the graph traversal in a way such that it maximizes the maximum similarity to the destination view along the path.

%In summary of this section, a procedure is proposed to recover a number of possible poses from a pose-graph in an efficient way. The proposed heuristic for the $\AStar$ algorithm overcomes the problem of not having a global scale when measuring the distance of two views. 

\section{Guided Matching with Pose}
\label{sec:pose-based-matching}
When matching an image collection, the most time-consuming process is often the local descriptor matching~\cite{IMC2020} \emph{aka} establishing tentative correspondences. The reason is that it has $\mathcal{O}(n^2)$ complexity both, \emph{w.r.t.}\ the number of local features, \ie\ for a single image pair, and \emph{w.r.t.}\ the number of images, \ie for the whole collection. We have already addressed the second problem (see Section~\ref{sec:notation}), but matching a single image pair still takes a significant amount of time. The common way to accelerate feature matching is by using approximate nearest neighbor search, instead of the exact one, \eg using the kd-tree algorithm as implemented in FLANN~\cite{FLANN2009}. Yet, even the approximate matching still takes a considerable amount of time and decreases the accuracy of the camera pose~\cite{IMC2020}.
We propose an alternative solution instead -- to exploit the poses coming from walks in the current pose-graph to establish tentative correspondences.
These poses will be used to make the standard descriptor matching ``light-weight'' by checking only those correspondences which are consistent with the pose. 
%two images, one of the  usually is a time-consuming procedure to establish tentative correspondences ...

\begin{figure}
  	\centering
  	\includegraphics[width=0.99\columnwidth]{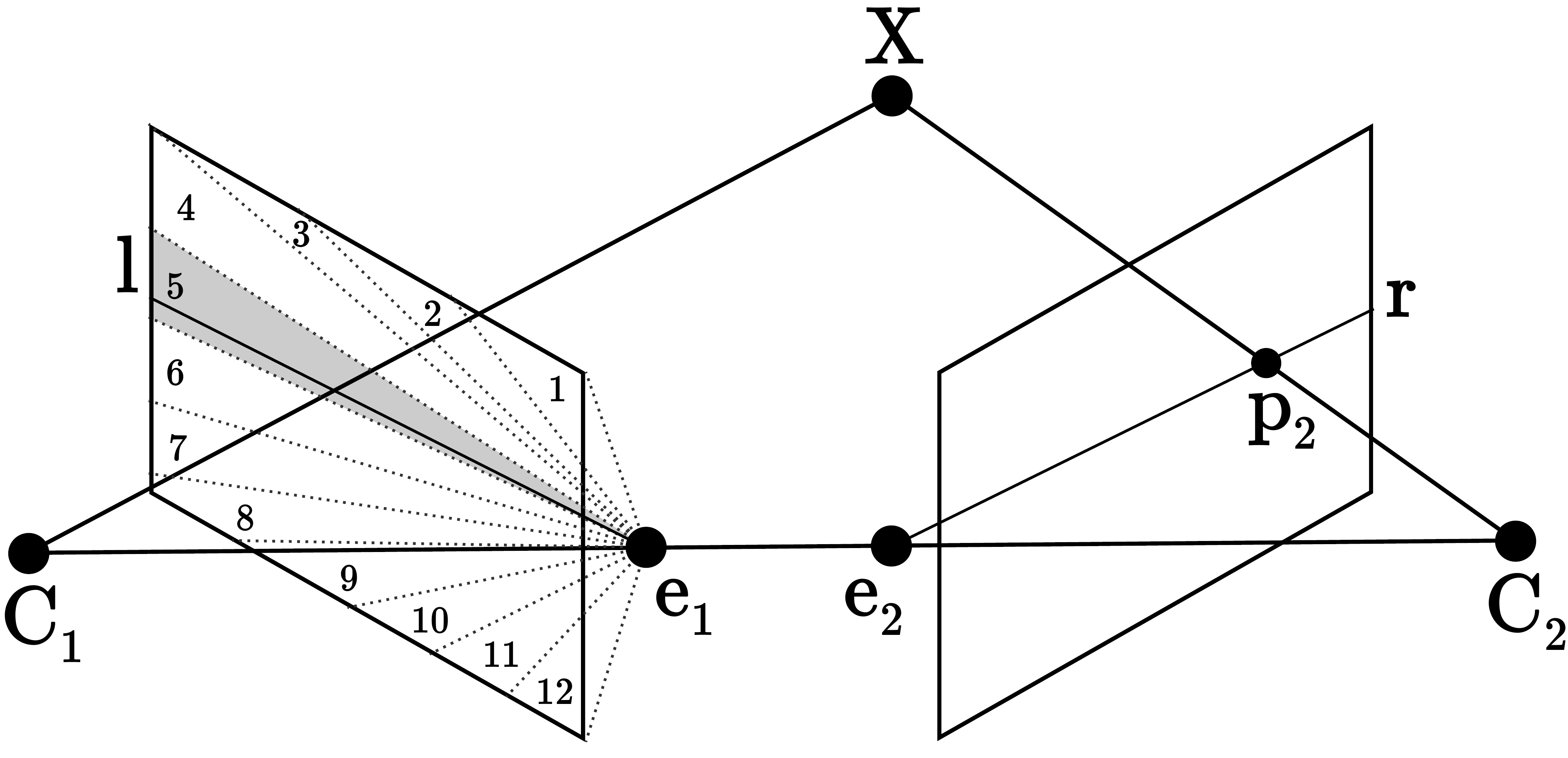}
    \caption{ \textit{Epipolar Hashing}. 
    Point $\mat p_2$ in image $\mat C_2$ is assigned to the bin in $\mat C_1$ (gray area) which its corresponding epipolar line $\mat l$ selects.
    The bins are defined on the angles of the epipolar lines in $\mat C_1$. The epipoles are $\mat e_1$ and $\mat e_2$.
    }
    \label{fig:hashing_illustration}
\end{figure}

\vspace{1mm}
\noindent
\textbf{Guided feature matching with pose.}
%
%In order to avoid the $k$-nearest-neighbor search in the high-dimensional feature descriptor space, we propose to exploit the poses coming from walks in the current pose-graph as described earlier.
Let us assume that we are given sets of keypoints $\mathcal{K}_i, \mathcal{K}_j$ in the $i$th and $j$th images, respectively, and a relative pose $\mat P_{ij} = (\mat t_{ij}, \mat R_{ij}) \in \SE$ from the $i$th image to the $j$th one.
One can easily calculate essential matrix $\mat E_{ij} = [ \mat t_{ij} ]_{\times} \mat R_{ij}$ and use it to measure the distance $\epsilon$ of point pairs via the Sampson Distance or the Symmetric Epipolar Error~\cite{hartley2003multiple}.
Therefore, the objective is to find pairs of points $(\mat p_i, \mat p_j)$, where $\mat p_i \in \mathcal{K}_i$, $\mat p_j \in \mathcal{K}_j$ and $\epsilon(\mat p_i, \mat p_j, \mat E_{ij})$ is smaller than the inlier-outlier threshold.
In contrast to the traditional approach, where the feature matching is defined over the high-dimensional descriptor vectors of all possible keypoints using the $L_2$ norm, we propose to select a small subset of candidate matches using the essential matrix. 
Consequently, the descriptor matching becomes significantly faster. 
%Moreover, in practice, the traditional descriptor matching requires to find the $2$ nearest neighbors for each point to perform the standard SIFT ratio test~\cite{SIFT2004, IMC2020}.
%By using a distance defined by the essential matrix, it is enough to find only the closest neighbor.

Due to doing the matching in 2D, the procedure can be done by hashing instead of a brute-force or approximated pair-wise process. 
Using the essential matrix, finding possible pairs of a point in the source image degrades to finding points in the destination one where the corresponding epipolar lines project to the correct position, {\ie}, onto the selected point in the source image.   
Therefore, the points in the destination image can be put into bins according to their epipolar lines in the source image. A straightforward choice is to define the bins on the angles of the epipolar lines as it is visualized in Fig.~\ref{fig:hashing_illustration}. We call this technique in the further sections \textit{Epipolar Hashing} (EH).
Note that EH is applicable even when the intrinsic camera parameters are unknown and, thus, we only have a fundamental matrix.

Let us denote the angle of the corresponding epipolar line $\mat l$ in the first image of point $(x, y)$ in the second image as $\alpha_{(x, y)} \in [0, \pi)$. 
Due to the nature of epipolar geometry, certain $\alpha_{(x, y)}$ angles are impossible. Therefore, we define the interval consisting of the valid angles and, thus, which we will cover by a number of bins as $[a, b]$, where 
\begin{eqnarray*}
    a & = & \min \left(\alpha_{(0, 0)}, \alpha_{(w_2, 0)}, \alpha_{(0, h_2)}, \alpha_{(w_2, h_2)} \right), \\
    b & = & \max \left(\alpha_{(0, 0)}, \alpha_{(w_2, 0)}, \alpha_{(0, h_2)}, \alpha_{(w_2, h_2)} \right).
\end{eqnarray*}
Point $(0, 0)$ is the top-left corner of the second image, $w_2$ is its width, and $h_2$ is its height. 
When hashing the points, the size of a bin will be $\frac{b - a}{\# bins}$.
This is an important step in practice since sometimes the epipole is far outside the image and, thus, the range of angles is $<1$. Without the adaptive bin size calculation, the algorithm does not speed up the matching in such cases. 
Note that $[a, b]$ is $[0, \pi)$ when the epipole falls inside the image.
When doing the traditional descriptor matching, we consider only those matches which are in the corresponding bin and has lower Sampson distance than the threshold used for determining the pose. 

%\todo{Daniel, please check the sentence below. I could also add the graph, used for the ratio test correction}
 After the guiding is performed, descriptor matching is done on \num{2} to \num{30} possible candidates instead of all keypoints. To further clean it up, we apply standard SIFT ratio test~\cite{SIFT2004, IMC2020} with adaptive ratio threshold, depending on number of nearest neighbors -- the smaller the pool, the stricter the ratio test is.
 Details are added to the supplementary material.
%Guiding with EH and Sampson distance filtering reduces nearest neighbor candidates from 8000 to 2..30 number of possible Suchs 

The matching process is applied after $\AStar$ if that finds a good pose. 
Since $\AStar$ requires a set of correspondences to determine if a pose is reasonably good, we use correspondences from those point tracks where the current images are visible. 
The multi-view tracklets are calculated and updated when a new image pair is matched successfully. Since both global and incremental SfM algorithms require point tracks, this step does not add to the final processing time. \vspace{-1mm}

\begin{table}[t]
	\caption{Run-time of pose estimation on \num{402130} view pairs from the 1DSfM dataset using GC-RANSAC, breadth-first traversal and $\AStar$ with the proposed heuristic (unit: seconds). }
	\vspace{-20pt}
	\begin{center}
		\setlength{\tabcolsep}{3.0mm}{
			\resizebox{0.99\columnwidth}{!}{
				\begin{tabular}{ c || c | c | c}
                    \hline
                    \rowcolor{black!10} 
                      \cellcolor{black!10}method & avg & med & total \\
                    \hline
                        \rowcolor{black!0} GC-RANSAC~\cite{barath2018graph} & 0.815 & 0.915 & \phantom{1 }327 574 \\ 
                        \rowcolor{black!0}Breadth-first & 2.916 & 0.703 & 1 429 423 \\
                        \rowcolor{black!5} 
                        $\AStar$ & \textbf{0.173} & \textbf{0.056} & \textbf{\phantom{1 3}82 672} \\
					\hline
		\end{tabular}}}
	\end{center}
	\label{tab:ransac_times}
\end{table}

\section{Adaptive Correspondence Ranking}

In this section, we propose a strategy to adaptively set the weight of the point correspondences for PROSAC sampling~\cite{chum2005matching} when doing pair-wise relative pose estimation in large-scale problems. 
PROSAC exploits an a priori predicted inlier probability rank of the points and starts the sampling with the most promising ones. 
Progressively, samples which are less likely to lead to the sought model are drawn. 
The main idea of the proposed algorithm is based on the fact that features detected in one image and matched to the other ones often appear multiple times when matching the image collection. 
Therefore, correspondences containing points which were inliers earlier are to be used first in the PROSAC sampling. Conversely, points that were outliers in the previous images should be drawn later.  

Assume that we are given the $t$-th image pair to match with sets of keypoints $\mathcal{K}_i$, $\mathcal{K}_j$. Each keypoint $\mat p$, from either set, has score $s_{\mat p}^{(t)} \in [0, 1]$ for determining its outlier rank among all keypoints. 
After successfully estimating the pose $\mat P_{ij}$ of the image pair, we are given the probability $P((\mat p, \mat q) \; | \; \mat P_{ij})$ of $(\mat p, \mat q)$ being outlier given pose $\mat P_{ij}$, where $(\mat p, \mat q)$ is a tentative correspondence, $\mat p \in \mathcal{K}_i$, $\mat q \in \mathcal{K}_j$. Probability $P((\mat p, \mat q) \; | \; \mat P_{ij})$ can be calculated, \eg, as in MSAC~\cite{torr2000mlesac}, MLESAC~\cite{torr2000mlesac} or MAGSAC++~\cite{barath2020magsacpp} from the point-to-model residuals assuming normal or $\chi^2$ distributions.
Since we do not know how probabilities $P(\mat p \; | \; \mat P_{ij})$ and $P(\mat q \; | \; \mat P_{ij})$ relate, we assume that $\mat p$ and $\mat q$ being consistent with the rigid reconstruction are independent events and, thus, $P((\mat p, \mat q) \; | \; \mat P_{ij}) = P(\mat p \; | \; \mat P_{ij}) P(\mat q \; | \; \mat P_{ij})$.
To be able to decompose probability $P((\mat p, \mat q) \; | \; \mat P_{ij})$, we assume that $P(\mat p \; | \; \mat P_{ij}) = P(\mat q \; | \; \mat P_{ij}) = \sqrt{P((\mat p, \mat q) \; | \; \mat P_{ij})}$.
This probability is then used to update score $s_{\mat p}$ and $s_{\mat q}$ after the $t$-th image pair matched as $s_{\mat p}^{(t + 1)} = s_{\mat p}^{(t)} P(\mat p \; | \; \mat P_{ij})$ and  $s_{\mat q}^{(t + 1)} = s_{\mat q}^{(t)} P(\mat q \; | \; \mat P_{ij})$.
Let us set $s_{\mat p}^{(0)} = 1$ since all keypoints are similarly likely to be outliers in the beginning.

\begin{table}[t]
	\caption{Run-time of matchers used for forming tentative correspondences with and without exploiting the relative pose in the proposed way (unit: seconds).}
	\vspace{-20pt}
	\begin{center}
		\setlength{\tabcolsep}{1.0mm}{
			\resizebox{0.99\columnwidth}{!}{
				\begin{tabular}{ c || c | c || c | c}
                    \hline
                      \rowcolor{black!10} 
                      & \multicolumn{2}{ c }{avg} & \multicolumn{2}{ c }{med} \\
                      \rowcolor{black!10} \cline{2-5}\rowcolor{black!10}
                      \multirow{-2}{*}{matcher} & w/o pose & pose & w/o pose & pose \\
                    \hline
                        \rowcolor{black!0}Brute-force & 7.609 & 1.078 & 1.047 & 1.139 \\ %10.77
                        \rowcolor{black!0}FLANN~\cite{FLANN2009} & 0.992 & 0.318 & 0.728 & 0.137 \\ % 5.48
                        \rowcolor{black!5} 
                        Epipolar Hashing & -- & \textbf{0.057} & -- & \textbf{0.046} \\
					\hline
		\end{tabular}}}
	\end{center}
	\label{tab:matching_times}
\end{table}

When the $(t+1)$th image pair is matched by using PROSAC sampling, the correspondences are ordered according to their outlier ranks $s_{\mat p}^{(0)}$ increasingly, such that the first one is the least likely to be an outlier.

\section{Experiments}

\begin{figure*}
  	\centering
  	\includegraphics[trim={1mm 0mm 2mm 0mm},clip,width=0.325\textwidth]{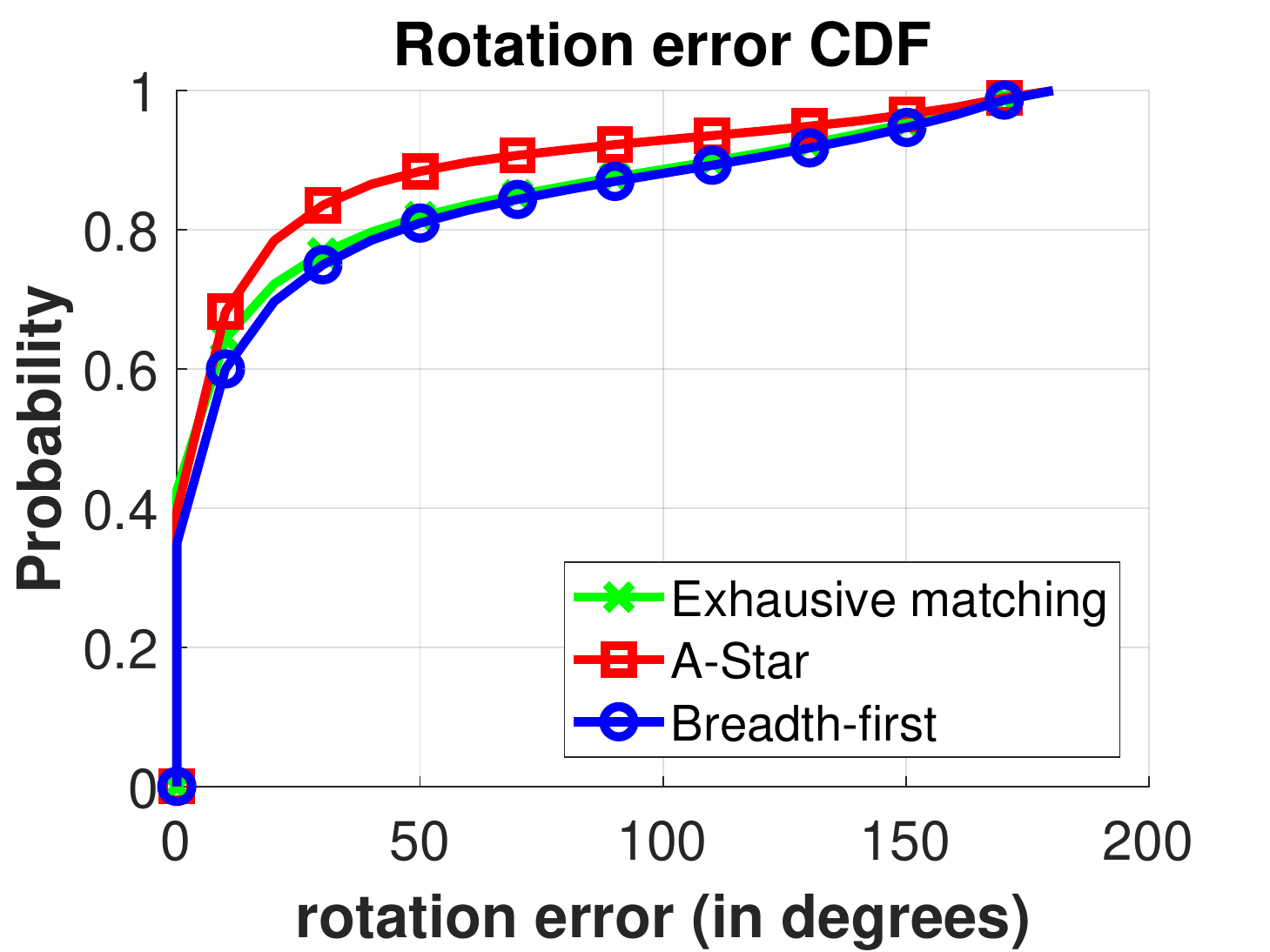}\hfill
  	\includegraphics[trim={1mm 0mm 2mm 0mm},clip,width=0.325\textwidth]{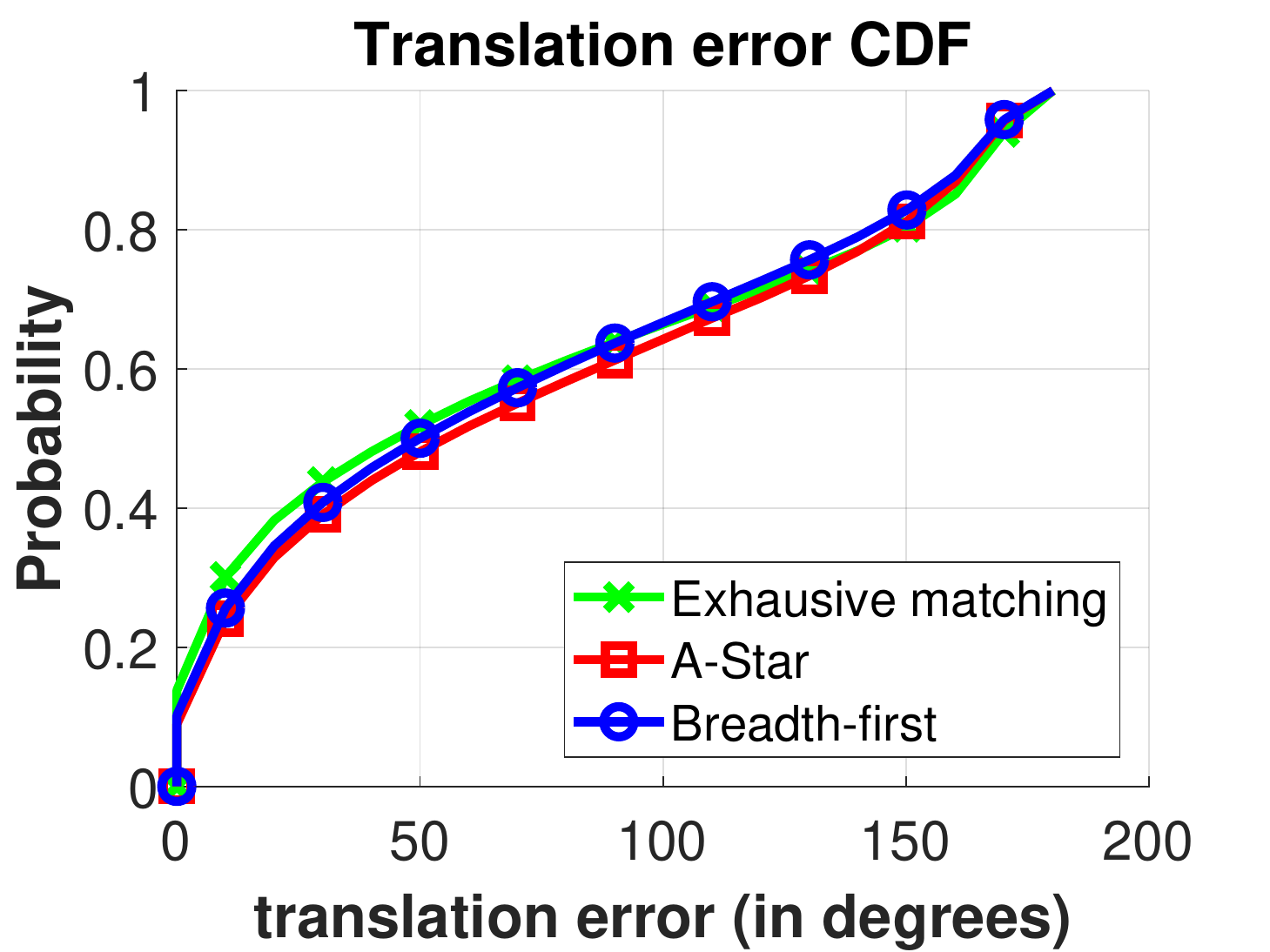}\hfill
  	\includegraphics[trim={1mm 0mm 2mm 0mm},clip,width=0.325\textwidth]{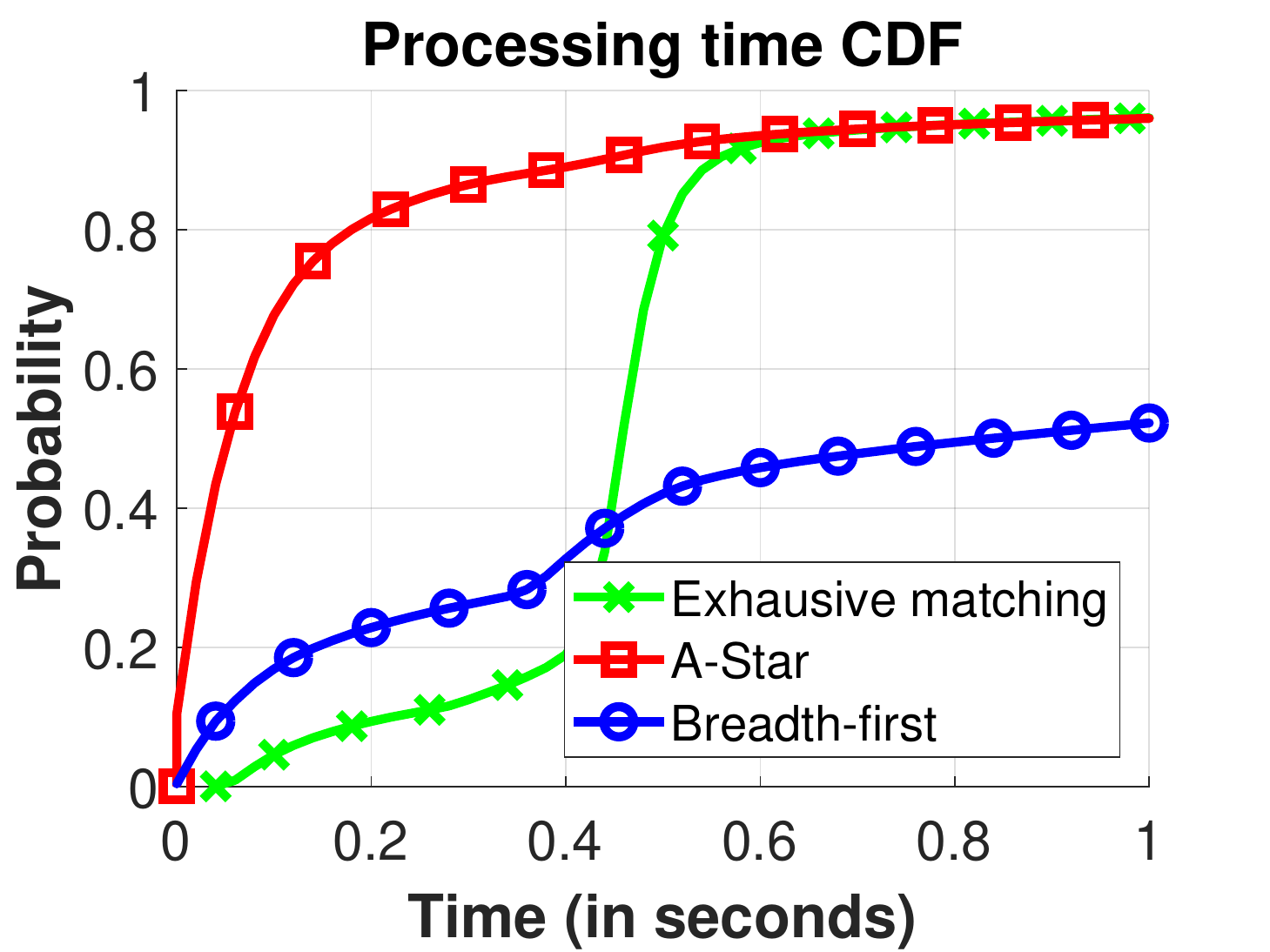}
    \caption{ The cumulative distribution functions of the errors (in degrees) and processing times (in seconds) of the pose-graphs generated by different initialization techniques on all scenes of the 1DSfM dataset. All algorithms returned \num{402130} poses.}
    \label{fig:traversal_cdf}
\end{figure*}

\begin{table*}[t]
\caption{The results of a global SfM~\cite{theia-manual} algorithm averaged over all scenes from the 1DSfM dataset~\cite{wilson_eccv2014_1dsfm}. The global SfM is initialized with pose-graphs generated by the traditional exhaustive matching (EM), breadth-first-based (BF), $\AStar$-based ($\AStar$), and Spanning-tree-based (MST) pose-graph building.
The reported properties are: the pose-graph generator method (1st), the number of views (2nd) and multi-view tracks (3rd) reconstructed by the global SfM procedure given an initial pose-graph; the total time of the pose-graph generation and bundle adjustment on a single CPU (4th); rotation error of the reconstructed global poses in degrees (5th), position error in meters (6th) and focal length errors (7th). }
\label{tab:1DSfM}
\resizebox{1.0\textwidth}{!}{%
\setlength\aboverulesep{0pt}\setlength\belowrulesep{0pt}%
\begin{tabular}{c | S[detect-weight,table-format=3.0] S[detect-weight,table-format=5.0] |  S[detect-weight,table-format=3.0] | S[detect-weight,table-format=2.1] S[detect-weight,table-format=2.1] S[detect-weight,table-format=2.1] | S[detect-weight,table-format=2.1] S[detect-weight,table-format=2.1] S[detect-weight,table-format=2.1] |S[detect-weight,table-format=1.1] S[detect-weight,table-format=1.1] S[detect-weight,table-format=1.1]}
\toprule
\rowcolor{black!10} 
 &
   &
   &
  {time} &
  \multicolumn{3}{c|}{\cellcolor{black!10}orientation err (\SI{}{\degree})} &
  \multicolumn{3}{c|}{\cellcolor{black!10}position err (m)} &
  \multicolumn{3}{c}{\cellcolor{black!10}focal len. err\ ($\times \num{e-2}$)} \\
\hline
\rowcolor{black!10}
{} & {$\#$ views} & {$\#$ tracks} & {(hours)} & {AVG} & {MED} & {STD}  & {AVG} & {MED} & {STD} & {AVG} & {MED} & {STD}   \\ \midrule
\rowcolor{white}
EM & 437 & 76837 & 202 & 9.8 & 6.6 & 6.2 & 10.9 & 8.4 & 13.9 & \bfseries 2.0 & \bfseries 1.2 & 3.7 \\ 
\rowcolor{white}
BF & \bfseries 463 & 73864 & 507 & \bfseries 7.7 & 7.0 & 7.1 & \bfseries 10.7 & \bfseries 6.1 & 18.8 & 2.2 & \bfseries 1.2 & 3.9 \\ 
\rowcolor{black!5}
\textbf{$\AStar$} + FLANN & 439 & 69312 & 58 & 8.6 & 6.9 & 6.6 & 11.2 & 6.6 & 19.4 & 2.2 & \bfseries 1.2 & 3.8 \\ 
\rowcolor{black!5}
\textbf{$\AStar$ + EH} & 444 & \bfseries 78335 & 29 & 7.8 & \bfseries 6.1 & \bfseries 5.2 & 10.8 & 6.2 & 19.5 & 2.4 & 1.4 & 4.1 \\ 
\rowcolor{white}
MST & 84 & 10334 & \bfseries 4 & 25.4 & 10.6 & 7.6 & 12.5 & 23.0 & \bfseries 7.9 & 2.4 & 1.5 & \bfseries 3.5 \\ \midrule
\bottomrule
\end{tabular}
}%
\end{table*}

%In this section, we show that the proposed techniques lead to a consistent and significant speedup with no noticeable deterioration in the accuracy of both the initial pose-graph and the final output of a state-of-the-art global SfM algorithm.

We tested the proposed algorithms on the 1DSfM dataset~\cite{wilson_eccv2014_1dsfm}.
It consists of \num{13} scenes of landmarks with photos of varying sizes collected from the internet. 
1DSfM provides 2-view matches with epipolar geometries and a reference reconstruction from incremental SfM (computed with Bundler~\cite{snavely2006photo,snavely2008modeling}) for measuring error. 
We used the SIFT features~\cite{SIFT2004} as implemented in OpenCV with RootSIFT~\cite{RootSIFT2012} descriptors. 
In each image, $8000$ keypoints are detected in order to have a reasonably dense point cloud reconstruction and precise pair-wise geometry camera poses~\cite{IMC2020}.
We combined mutual nearest neighbor check with standard distance ratio test~\cite{lowe1999object} to establish tentative point correspondences, as recommended in~\cite{IMC2020}.
The bin number for Epipolar Hashing was set to \num{45}. 
We matched all image pairs with global similarity higher than \num{0.4} with which we got accurate reconstruction in reasonable time.
The algorithms were tested on a total of \num{402130} image pairs.

The methods are implemented in C++ using the Eigen and Sophus~\cite{sophus-manual} libraries. The graph traversal algorithms are implemented by ourselves. For robust estimation, we always use the GC-RANSAC algorithm~\cite{barath2018graph} with the five-point algorithm of Stewenius {\etal}~\cite{stewenius2006recent}.
Note that the used GC-RANSAC implementation contains PROSAC sampling~\cite{chum2005matching}, SPRT test~\cite{chum2008optimal} and a number of sample and model degeneracy tests to be as efficient as possible.

\subsection{Alternatives for RANSAC}

Pose-graph generation algorithms are compared in this section, including the proposed $\AStar$-based technique. 
The compared methods are: \\[2mm]
\noindent
1.~The standard exhaustive matching (EM) where each tested image pair is matched.\\[1mm]
\noindent
2.~A minimal spanning tree (MST) where the global similarity score is used as weights.\\[1mm]
\noindent
3.~The proposed $\AStar$-based technique, where the pose comes from a path determined by $\AStar$ if possible. Otherwise, the standard matching is applied. \\[1mm]
\noindent
4.~Breadth-first (BF) traversal applied similarly as the proposed $\AStar$ algorithm.\\[2mm]
%\begin{enumerate}
%    \item The standard exhaustive matching (EM) where each tested image pair is matched %by GC-RANSAC.
%    \item The proposed $\AStar$-based technique, where the pose comes from a path %determined by $\AStar$ if possible. Otherwise, the RANSAC-based matching is %applied.
%    \item Breadth-first (BF) traversal applied similarly as the proposed $\AStar$ %algorithm.
%    \item A minimal spanning tree (MST) where the global similarity score is used as %weights.
%\end{enumerate}
%
The cumulative distribution functions of the rotation and translation errors (in angles) and processing times (in seconds) are shown in Fig.~\ref{fig:traversal_cdf}. 
We do not include MST here since it matches significantly fewer image pairs (\num{9922}) than the other methods (\num{402130}). 
In terms of accuracy, the $\AStar$-based technique leads to the most accurate rotation matrices while having similar translation errors as the breadth-first-based and exhaustive matching.
%Even though the rotation error of the minimal spanning tree seems extremely low, let us point to the fact that it estimates the pose only for a small percentage of the view pairs. 
%As it will be shown slightly later, this low number of view pairs severely affects the quality of the reconstruction by global SfM.
In terms of processing time, the proposed $\AStar$-based technique leads to a significantly more efficient pose-graph generation than EM or BF.
Note that the break-points in the processing time curves are caused by setting the maximum number of RANSAC iterations to \num{5000}. This is to avoid running RANSAC for extremely long when the inlier ratio is low. 

The average, median and total processing times of the pair-wise pose estimation algorithms are shown in Table~\ref{tab:ransac_times}. The run-times contain those cases as well when no valid pose was found by the $\AStar$ or breadth-first traversals and, thus, GC-RANSAC was applied to recover the pose. The $\AStar$ algorithm leads to a speedup of almost an order of magnitude with its median time being approx.\ $20$ times lower than that of GC-RANSAC. 
It validates the proposed heuristic that the breadth-first algorithm is significantly slower than $\AStar$.
Consequently, the proposed heuristic guides the path-finding in the pose-graph successfully.

\subsection{Matching with Pose}

We compare the feature matching speed with or without exploiting the pose determined by the $\AStar$ algorithm. 
In FLANN and brute-force matching, this means that we find all candidate matches which lead to smaller epipolar error than the inlier-outlier threshold. The best candidate is then selected by descriptor-based matching.
%Also, the proposed Epipolar Hashing is compared with brute-force and FLANN-based strategies. 
The run-times are reported in Table~\ref{tab:matching_times}.
Using the pose speeds up both the FLANN-based and brute-force algorithm significantly. 
The proposed Epipolar Hashing leads to a more than \num{17} times speedup compared to the traditional FLANN-based feature matching.
Also, by the Epipolar Hashing, the neighbors are found precisely without approximation as done in FLANN.

%Also, the neighbors are found precisely and not approximately as in FLANN.

\subsection{Adaptive Ranking}

The average, median and total processing times (in seconds) of the robust estimation using different correspondence ranking techniques for PROSAC are shown in Table~\ref{tab:prosac_times}. 
Three methods are compared: the uniform matching proposed originally for RANSAC (unordered); PROSAC when the correspondences are ordered according to their SIFT ratios~\cite{chum2005matching}; and the proposed adaptive re-ranking considering the prior information about the points from earlier estimations. 
While ordering the correspondences according to their SIFT ratios speeds up the estimation by \SI{10}{\percent} compared to the uniform sampling, the proposed adaptive re-ranking leads to an additional \SI{8}{\percent} speedup on average. 

\begin{table}[t]
	\caption{Run-time of PROSAC using different ordering techniques: unordered (RANSAC-like uniform~\cite{fischler1981random}), SIFT ratio~\cite{chum2005matching} and the proposed adaptive one (unit: seconds). }
	\vspace{-20pt}
	\begin{center}
		\setlength{\tabcolsep}{3.0mm}{
			\resizebox{0.99\columnwidth}{!}{
				\begin{tabular}{ c || c | c | c}
                    \hline
                    \rowcolor{black!10} 
                      \cellcolor{black!10}ordering & avg & med & total \\
                    \hline
                        \rowcolor{white}
                        Unordered & 0.736 & 0.768 & 295 997  \\ 
                        \rowcolor{white}
                        SIFT ratio & 0.664 & 0.698 & 266 849 \\
                        \rowcolor{black!5} 
                        Adaptive ordering & \textbf{0.615} & \textbf{0.643} & \textbf{247 320} \\
					\hline
		\end{tabular}}}
	\end{center}
	\label{tab:prosac_times}
\end{table}

\subsection{Applying Global SfM Algorithm}

Once relative poses are estimated for camera pairs of a given dataset, along with the inlier correspondences, they are fed to the Theia library~\cite{theia-manual} that performs global SfM~\cite{chatterjee2013efficient,wilson_eccv2014_1dsfm} using its internal implementation.
That is, feature extraction, image matching and relative pose estimation were performed by our code either using the proposed algorithm or the traditional brute-force pair-wise matching using the 5PT~\cite{stewenius2006recent} solver.
The key steps of global SfM are robust orientation estimation, proposed by Chatterjee {\etal}~\cite{chatterjee2013efficient}, followed by robust nonlinear position optimization by the method of Wilson {\etal}~\cite{wilson_eccv2014_1dsfm}.
The estimation of global rotations and positions enables triangulating 3D points, and the reconstruction is finalized by the bundle adjustment of camera parameters and point coordinates.
Since the reconstruction always failed on scene Gendarmenmarkt, we did not consider that scene when calculating the errors.

Table~\ref{tab:1DSfM} reports the results of Theia initialized by pose-graphs generated by the traditional exhaustive matching (EM), breadth-first graph traversal (BF), the proposed $\AStar$-based graph-traversal, and by using a minimum spanning tree (MST). 
While the generation of the minimum spanning tree-based pose-graph is extremely fast, it can be seen that it is not good enough for the global SfM algorithm to provide a reconstruction of reasonable size. The average number of views reconstructed when initialized by MST is significantly lower than using the other techniques.
It can be seen that the proposed $\AStar$-based methods lead to similar number of views and similar error to the traditional approach.
%Note that breadth-first traversal leads to the most views reconstructed, though, it is extremely slow.

\subsection{Heuristic for A\textsuperscript{*} Traversal}

In this section, the $\AStar$ traversal parameters are tuned on scene Alamo. 
For this purpose, the ground truth pose-graph is loaded and the pose is obtained by $\AStar$ between all possible image pairs which were not connected directly in the graph. 
The parameters tuned are the weight $\lambda$ from \eqref{eq:heuristics} and the maximum depth allowed when obtaining the walks. 
%The parameters which perform the best are tuned only on Alamo.
They were then used for all other tests and scenes.

\begin{figure}
  	\centering
  	\includegraphics[width=0.49\columnwidth]{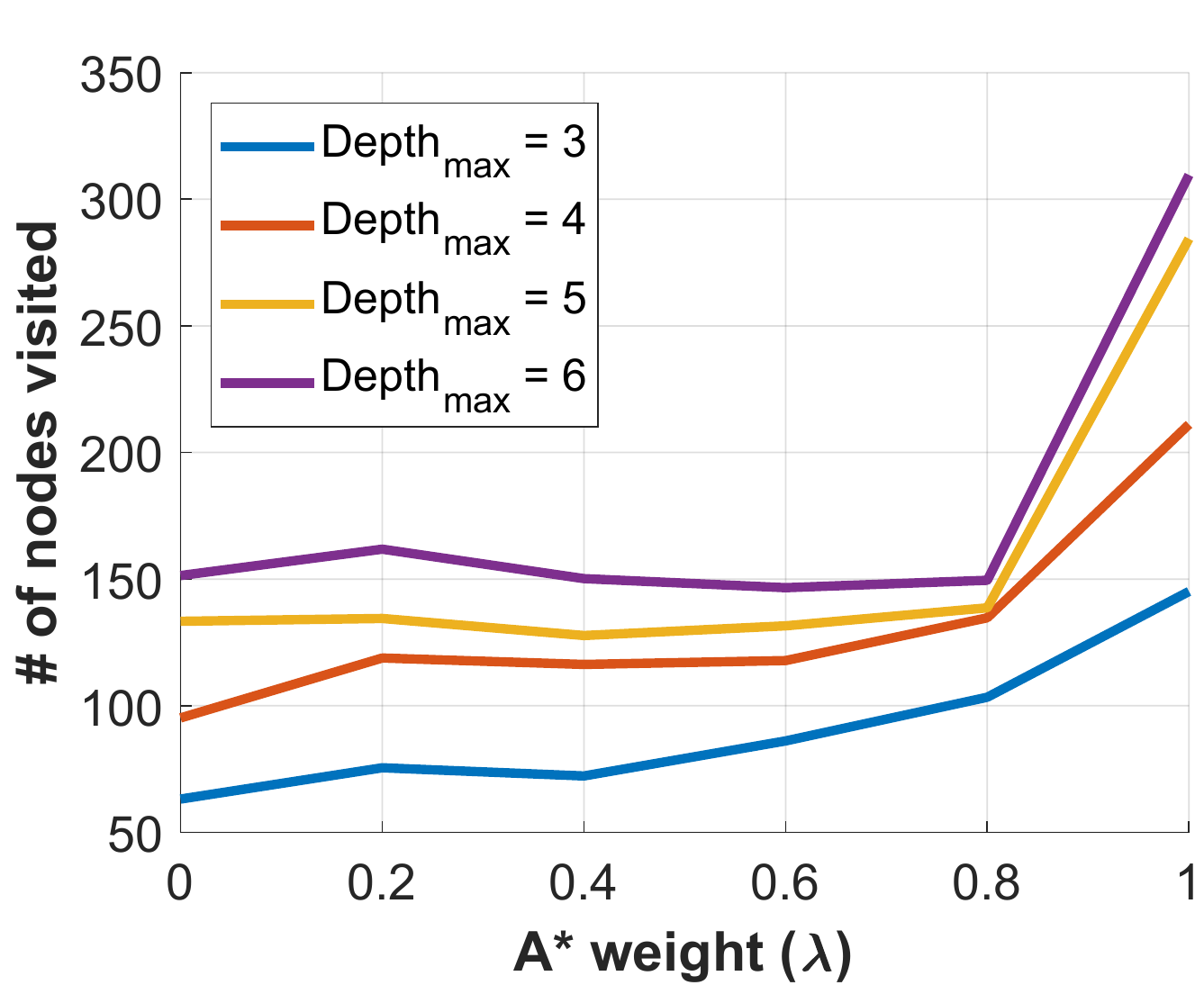}
  	\includegraphics[width=0.49\columnwidth]{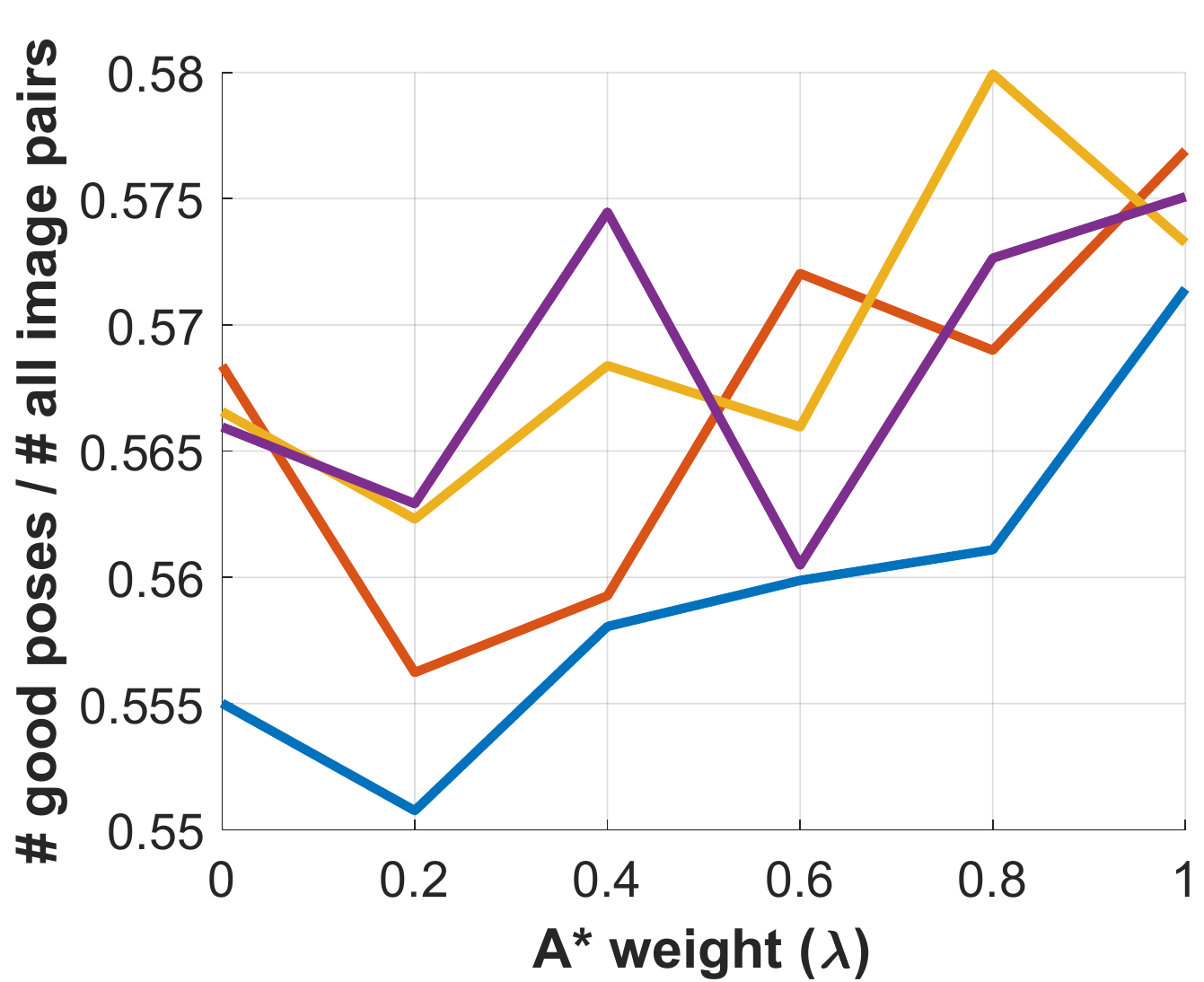}
    \caption{ Average \# of nodes visited by $\AStar$ (left) and ratio of accurate relative poses obtained (right) in scene Alamo plotted as the function of weight $\lambda$ and the max.\ depth. }
    \label{fig:heuristics_experiment}
\end{figure}

In Fig.~\ref{fig:heuristics_experiment}, the average number of nodes visited by the $\AStar$ traversal (left) and the ratio of accurate relative poses obtained (right) are plotted as the function of weight $\lambda$. 
Different maximum depths are shown by the different curves. 
Parameter $\lambda = 0$ means that there is no constraint on the edge quality, the only goal is to get to a node similar to the destination. 
Parameter $\lambda = 1$ is interpreted as walking on the highest quality edges without trying to get close to the destination.
The number of nodes visited, \ie proportional to the processing time, is nearly constant for $\lambda \in [0, 0.8]$. The highest success rate is achieved by $\lambda = 0.8$ and $\text{Depth}_{\text{max}} = 5$. Since $\lambda = 0.8$ with $\text{Depth}_{\text{max}} = 5$ also leads to a reasonably low number of nodes visited, we chose these values in all our experiments. 

\section{Conclusions}

The final bundle adjustment of global SfM algorithms has a negligible time demand compared to the initial pose-graph generation.
To speed this step up by almost an order of magnitude, we proposed three new algorithms. 
The standard procedure ({\ie}, feature matching by FLANN; pose estimation by RANSAC-like robust estimation) for estimating the pose-graph for all scenes from the 1DSfM dataset took a total of \num{726487} seconds on a single CPU -- approx.\ \num{202} hours. 
By using the proposed set of algorithms ({\ie}, $\AStar$-based pose estimation; Epipolar Hashing for matching; adaptive re-ranking), the total run-time is reduced to \num{105593} seconds (\num{29} hours).
In the experiments, $\AStar$ found a valid pose in \SI{93.8}{\percent} of the image pairs. Thus, traditional FLANN-based feature matching and pose estimation by RANSAC was applied only to \SI{6.2}{\percent} of the image pairs.

%Finally, a global bundle adjustment obtains the accurate reconstruction from the pair-wise poses. Interestingly, this step has negligible time demand, \ie, less than a minute in our experiments, compared to the initial pose-graph generation.

% The theoretical lower bound of the number of cases when the traditional matching have to applied is $n - 1$, \ie, to generate a spanning tree. Its upper bound is $\binom{n}{2}$. 
{\small
\bibliographystyle{ieee_fullname}
\bibliography{egbib}
}

\clearpage
%%%%%%%%% BODY TEXT
\section{Second Nearest Ratio Test with Pool Size-sensitive Threshold}
\label{sec:intro}

This section supplement Section 3 in the main paper. 
The common way of filtering unreliable tentative correspondence is the second-nearest ratio test (\emph{aka} SIFT ratio test or Lowe ratio test)~\cite{SIFT2004, IMC2020}.
In this test, after the descriptor matching, tentative correspondences get rejected if their ``best'' match is not significantly closer than the second best one. 
Thus, correspondences are filtered if 
\begin{equation*}
    {\texttt{distance of 1st nearest neighbor} \over \texttt{distance of 2nd nearest neighbor}} > \gamma,
\end{equation*}
where $\gamma$ is the SIFT ratio threshold. Parameter $\gamma$ is typically set to $0.7-0.9$, the common default being 0.8.

When applying the proposed \textit{Epipolar Hashing} algorithm to select a subset of candidate matches for each feature point, this SIFT ratio test is rendered almost completely ineffective without the adaptation of threshold $\gamma$.
This is caused by the fact that \textit{Epipolar Hashing} reduces the number of features in the pool from which the neighbors are selected, significantly, to $2-30$ on average in our experiments.
Due to this small pool, the density of points and thus the distance to second nearest descriptor is increased. Therefore the second best one is unlikely to be almost as close as the best match.
In such cases, the standard SIFT ratio test fails to filter incorrect correspondences. In other words, there are many false positive matches.

Let us assume that non-matching descriptors are randomly distributed \emph{w.r.t} the query descriptor. 
Consequently, the more descriptors we have in the pool, the lower
the distance to the closest ones to the query will be. 
Therefore, if an equally strict condition on the quality of the tentative correspondences is required, in terms of false positives, regardless the number of features detected, we need to adapt the SNN ratio test threshold $\gamma$ based on the number of features in the pool. 

\begin{figure}[htb]
  	\centering
  	\includegraphics[width=0.80\linewidth]{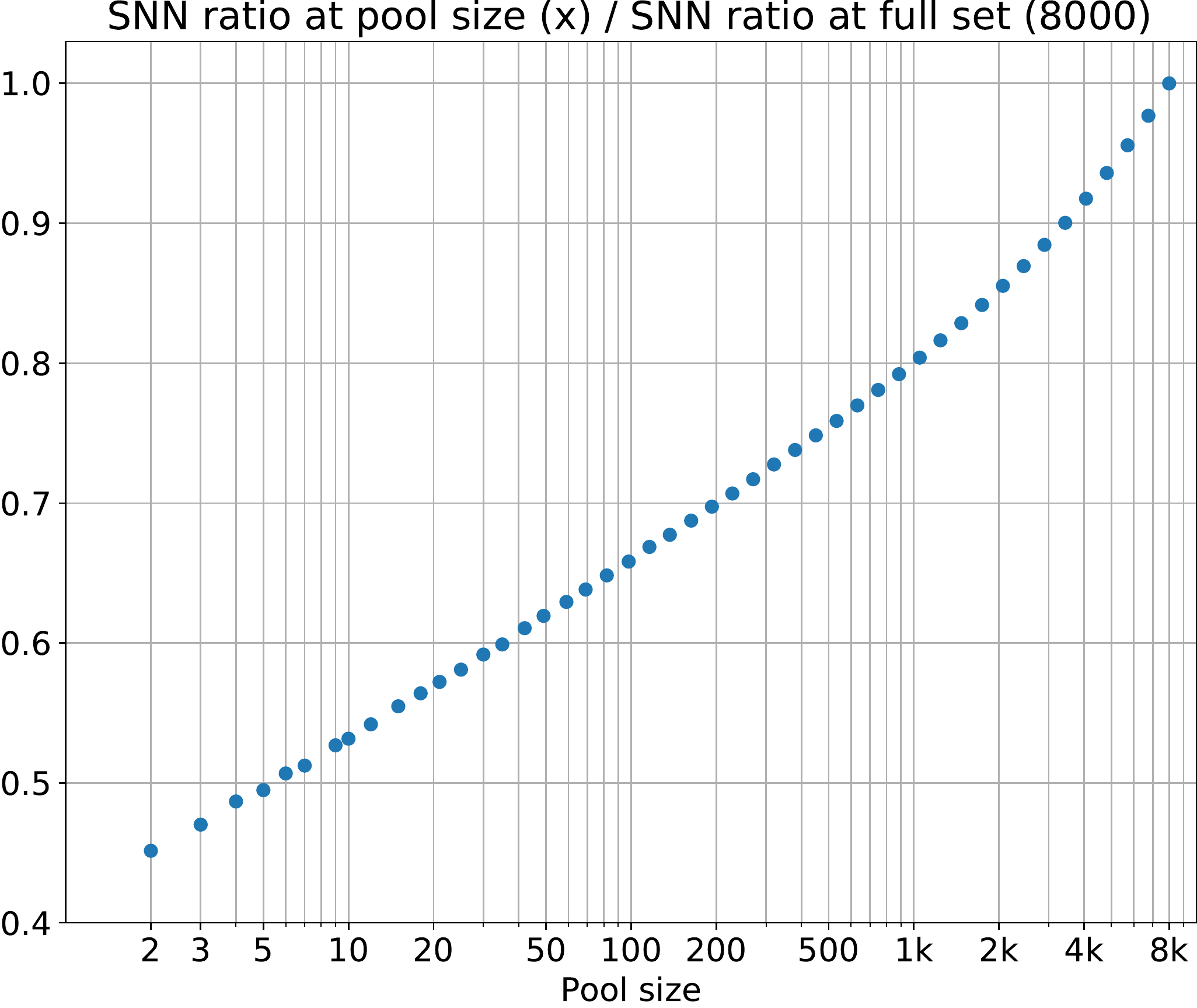}
  	\caption{Dependence of Lowe's SNN ratio on the the descriptor pool size. Averaged over HPatches image pairs, 8000 SIFT features.}
    \label{fig:adaptive-snn-ratio}
\end{figure}

\begin{table}[htb]
\caption{The results of a global SfM~\cite{theia-manual} algorithm on scene Madrid Metropolis with and without adaptive second nearest distance ratio when applying the proposed Epipolar Hashing.
The reported properties are: the number of views (2nd) and multi-view tracks (3rd) reconstructed by the global SfM procedure. }
\label{tab:1DSfM}
\centering
\resizebox{0.85\columnwidth}{!}{%
\setlength\aboverulesep{0pt}\setlength\belowrulesep{0pt}%
\begin{tabular}{c | S[detect-weight,table-format=3.0] S[detect-weight,table-format=5.0] }
\toprule
\rowcolor{black!10}
{} & {$\#$ views} & {$\#$ tracks}  \\ \midrule
\rowcolor{white}
w/o adaptive ratio test & 136 & 9486 \\ 
\rowcolor{white}
with adaptive ratio test & \bfseries 282 & \bfseries 29665  \\  \midrule
\bottomrule
\end{tabular}
}%
\end{table}

\begin{figure*}[htb]
  	\centering
  	\includegraphics[width=0.90\textwidth]{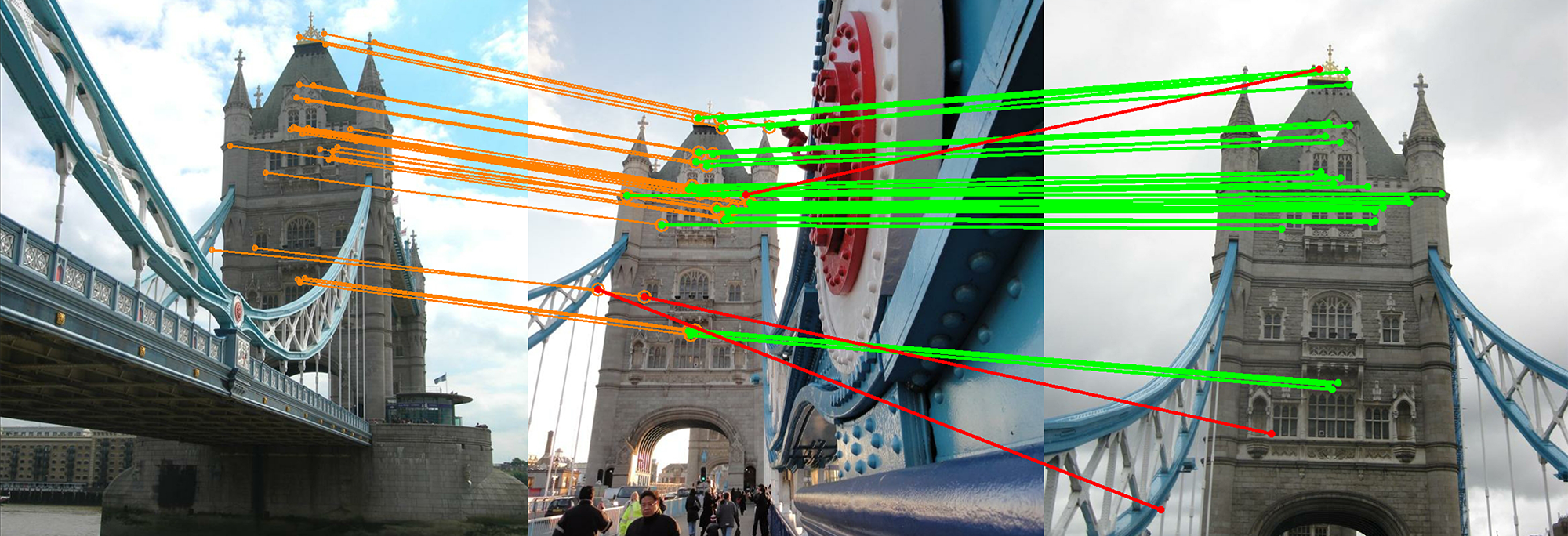}  	\includegraphics[width=0.90\textwidth]{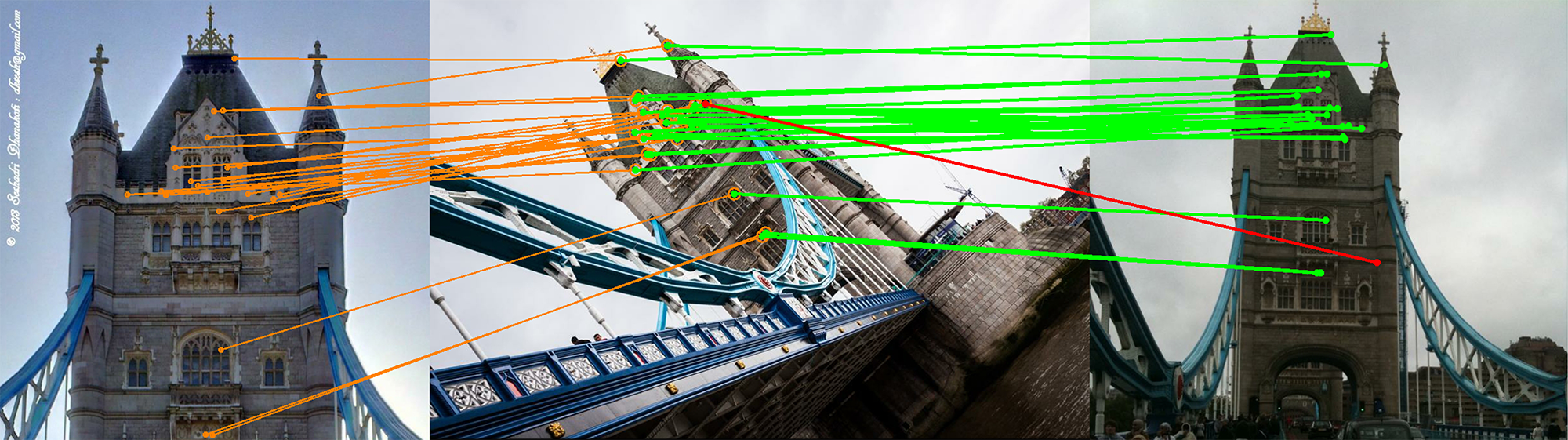}
  	\caption{Example triplets of images and the found inlier correspondences used for calculating the values in Fig.~\ref{fig:prosac}. (\textbf{Orange}) Inlier correspondences between the 1st and 2nd images which are visible in the 3rd one. (\textbf{Green}) Correspondences which got good rank by the proposed method and are consistent with the ground truth epipolar geometry between the 2nd and 3rd images. (\textbf{Red}) Correspondences which got good rank and are inconsistent with the epipolar geometry.
  	  Significantly more ``good'' correspondences got good ranking than incorrect ones -- the number of green points is higher than that of the red ones. }
    \label{fig:triplet}
\end{figure*}

\begin{figure}[htb]
  	\centering
  	\includegraphics[width=0.90\linewidth]{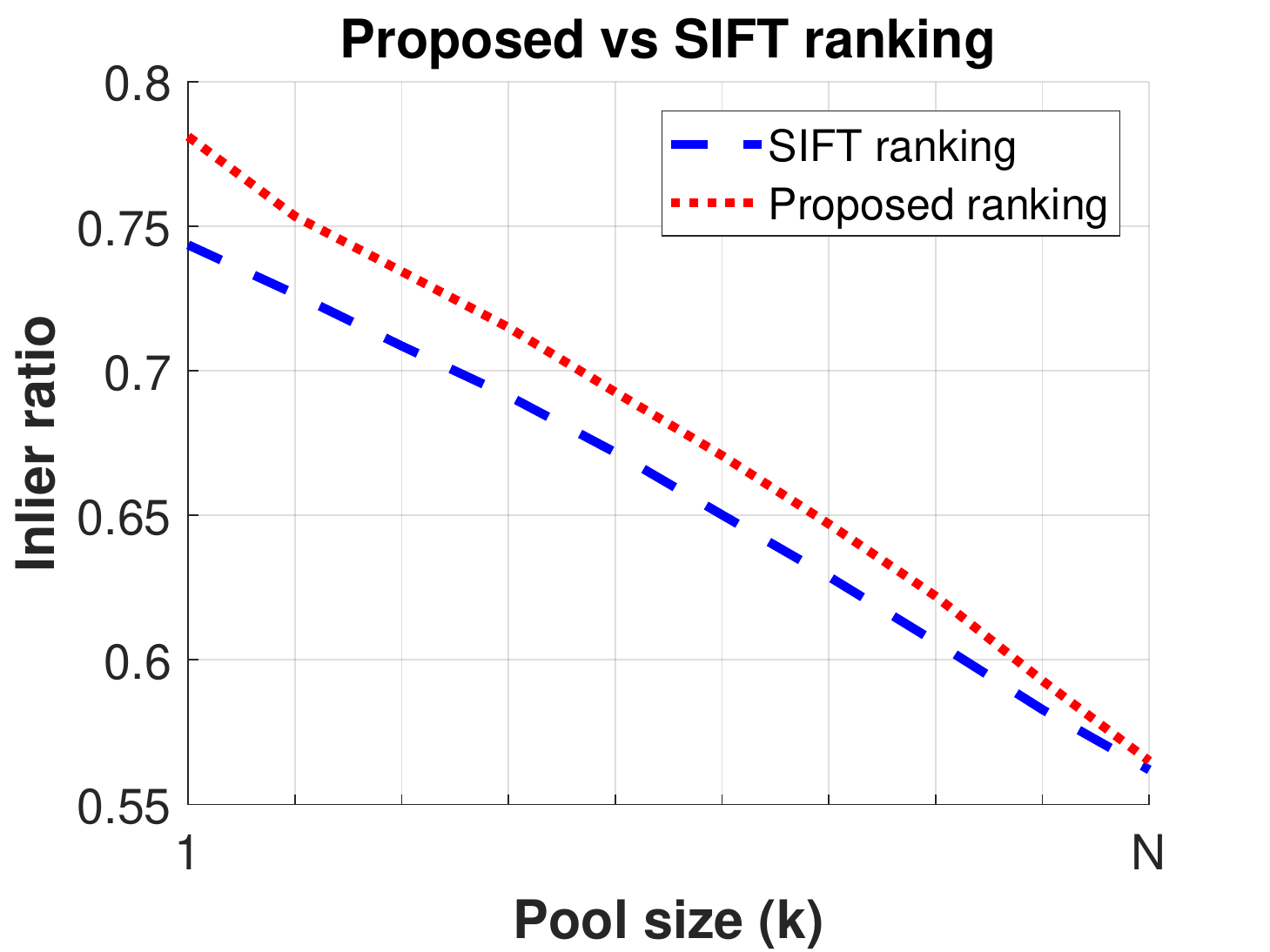}
  	\caption{Inlier ratio in the first $k$ correspondences when ordered by the proposed or SIFT rankings. The values are calculated from $500$ randomly selected image triplets from the London Bridge dataset.}
    \label{fig:prosac}
\end{figure}

The following experiment was run on each image pair from the HPatches-Sequences~\cite{hpatches2017} dataset.
First, 8000 SIFT features were detected in both images.
For each feature point, the nearest neighbor (minimizing the SIFT descriptor distance) and "reference" second nearest neighbor (second nearest at 8000) were found using the full set of features in the other image. 
The second nearest neighbor was selected from a random $p$-sized subset of points (second nearest at x). We then calculated the distance ratio of the nearest and second nearest neighbors. The results were averaged over all features and image pairs. 
In Fig.~\ref{fig:adaptive-snn-ratio}, this ratio is plotted as a function of the pool size $p$ from which the second nearest neighbor is selected.
The dependence of the SNN ratio on the feature pool size is almost linear in the log space. 
We use this dependence to correct the SNN ratio threshold -- the default value of $0.9$ for mutual SNN ratio~
\cite{IMC2020} is multiplied by the $y$-value depending on the feature number in the pool. For example, for 5 features, the resulting threshold is $0.45$.

Example reconstruction results with and without adaptive ratio test on scene Madrid Metropolis are shown in Table~\ref{tab:1DSfM}.
It can be seen that the adaptive ratio test is extremely important in Epipolar Hashing. The large difference in the number of reconstructed views is caused by the following phenomenon. The proposed $\AStar$-based algorithm first attempts to efficiently connect a new image to the pose-graph by extending tracks using Epipolar Hashing. If this process produces seemingly sufficient number of correspondences, full descriptor-based matching does not take place. A high number of false positives in the Epipolar Hashing process leads to an incorrect decision that full matching is not needed and an incorrect pose is obtained from the false positive matches, which are all, by construction, consistent with the initial estimated epipolar geometry, which is incorrect or very imprecise.
% Without it, Epipolar Hashing tends to return correspondences which are consistent with the 

\section{Adaptive Correspondence Ranking}

This section supplement Section 4 in the main paper. 
In order to compare the effect of the proposed correspondence re-ranking strategy, we selected $500$ image triplets from the London Bridge dataset, see Fig.~\ref{fig:triplet} for examples.
The images in each triplet were selected randomly but in a way to ensure that they have a commonly visible area. 
For each triplet, the epipolar geometry was estimated between the first two images by standard RANSAC.
Next, for estimating the relative pose between the second and third images, the correspondences were ordered either by the proposed re-ranking strategy or by their SIFT scores.
Finally, we measured the inlier ratio in the sets consisting of the first $k$ correspondences, $k \in [1, N]$.
Fig.~\ref{fig:prosac} plots the inlier ratio, averaged over the $500$ tests, as a function of the pool size $k$.
The proposed algorithm leads to a better ordering than exploiting the SIFT scores -- its inlier ratio is higher among the first $k$ correspondences.

\end{document}